\documentclass[10pt,twocolumn,letterpaper]{article}

\usepackage{cvpr}              
\definecolor{cvprblue}{rgb}{0.21,0.49,0.74}
\usepackage[pagebackref,breaklinks,colorlinks,allcolors=cvprblue]{hyperref}
\usepackage{newfloat}
\usepackage{listings}
\usepackage{algorithm}
\usepackage{algorithmic}
\usepackage{amsmath}
\usepackage{amssymb}
\usepackage{ftnxtra}
\usepackage{float}
\def\paperID{11757} 
\def\confName{CVPR}
\def\confYear{2026}

\title{Differentially Private 2D Human Pose Estimation}

\author{
Kaushik Bhargav Sivangi \quad
Paul Henderson \quad
Fani Deligianni \\
School of Computing Science, University of Glasgow \\
\texttt{\{kaushik.sivangi, paul.henderson, fani.deligianni\}@glasgow.ac.uk}
}

\begin{document}
\maketitle
\begin{abstract}
Human pose estimation (HPE) underpins critical applications in healthcare, activity recognition, and human-computer interaction. However, the privacy implications of processing sensitive visual data present significant deployment barriers in critical domains. 
Differential Privacy (DP) provides formal guarantees but often results in steep performance costs.
We introduce the first unified framework for differentially private  2D Human Pose Estimation (2D-HPE) that achieves strong privacy-utility trade-offs for structured visual prediction through complementary noise mitigation mechanisms. Our Feature-Projective DP integrates: (1) subspace projection that reduces noise variance by a factor $k/p$ by restricting gradient updates to a $k$-principal subspace within the full $p$-dimensional parameter space, and (2) feature-level privacy, which selectively privatizes sensitive features while retaining public visual cues. Together these mechanisms yield a multiplicative utility gain under formal privacy constraints.
Extensive experiments on MPII and HumanART datasets across privacy budgets $(\varepsilon \in \{0.2, 0.4, 0.6,  0.8\})$, clipping thresholds $(C \in \ \{0.01, 0.1, 1.0\})$ and training strategies demonstrate consistent improvements over vanilla DP-SGD. 
At $\varepsilon=0.8$, our method achieves 82.61\% PCKh@0.5, recovering 73\% of the privacy induced performance gap. Cross-dataset evaluation on the HumanART confirms generalization (51.6 AP). Our study provides the first rigorous benchmark and a practical blueprint for privacy-preserving pose estimation in sensitive, real-world applications. Project page: \textcolor{red}{https://bhairava2898.github.io/DP2DHPE/}
\end{abstract}    
\section{Introduction}
Human pose estimation (HPE) transforms raw visual data into structured keypoint representations of human posture and movements. This fundamental computer vision task enables numerous high impact applications in healthcare, activity recognition, human-computer interaction, sports analysis, and video games \cite{artacho2020unipose, mao2022poseur, Wang_2022_CVPR, Lu_2024_CVPR,li2022simcc,chharia2025mvssm}.
However, as these systems are increasingly deployed into sensitive environments such as hospitals, homes, and workplaces, they introduce severe privacy risks at multiple levels \cite{martinez2021ethical}.


At the data level, raw images contain identifiable biometric information exposed during collection and processing \cite{zheng2023deep}. At the model level, trained networks can inadvertently memorize training data, enabling adversaries to extract sensitive information through model inversion, membership inference and reconstruction attacks \cite{geiping2020inverting,Jegorova2023,zakariyya2025differentially}. 
For instance, an adversary can exploit a model's weights \cite{tian2025simudy,haim2022reconstructing} or gradients \cite{hatamizadeh2023gradient} to reconstruct distinctive physical characteristics of patients or sensitive contextual information, such as the patient's home environment from the private training dataset \cite{Jegorova2023}. This reconstruction could potentially identify individuals with specific medical conditions and reveal that they received treatment at a particular facility during the model's training period, thereby compromising both medical confidentiality and location privacy. 

Previous privacy-preservation approaches in HPE rely primarily on data anonymization techniques, such as blurring, pixelation, and template-based shape modeling ~\cite{ahmad2024event,Ruiz-Zafra2023,Hesse_2018_ECCV_Workshops}. While these methods provide some level of privacy protection, they are often task-specific and can severely compromise the utility of the data for broader analysis. For instance, anonymization that removes facial features might preserve basic joint position information but destroy crucial clinical indicators needed for stress level assessment or abnormal motion pattern detection ~\cite{barattin2023attribute}. Furthermore, these methods do not offer formal privacy guarantees and remain vulnerable to more sophisticated attacks, limiting their applicability in highly sensitive contexts ~\cite{zakariyya2025differentially,NEURIPS2022_OnionEffect}. Moreover, these approaches do not address the inherent vulnerability of neural networks to memorization attacks that can reconstruct training data \cite{haim2022reconstructing,tian2025simudy}, limiting model sharing for research and clinical deployment. 
The inherent tension between improving model utility and ensuring robust privacy preservation represents a challenging research problem \cite{Abbasi2025,fang2023improved,zhou2021bypassing} that has not been adequately explored in the context of HPE.

Differential privacy (DP) provides a principled framework for mitigating these risks by offering provable guarantees against information leakage from both data and model parameters \cite{hardt2010geometry,dwork2021promise, abadi2016deep,Dwork2006_DP}. 
However, implementing DP through DP Stochastic Gradient Descent (DP-SGD)~\cite{abadi2016deep}, typically results in substantial performance degradation, which is particularly problematic for fine-grained vision tasks like HPE where spatial precision is paramount ~\cite{yu2019differentially,de2022unlocking,Duan2025_DPGeometrically}. 

In this work, we present the first systematic framework for differentially private learning in 2D Human Pose Estimation. We demonstrate that directly applying DP-SGD to 2D-HPE models leads to significant degradation in utility due to the fine-grained nature of keypoint prediction.
We address the privacy-utility trade-off in 2D-HPE, through two complementary mechanisms. First, we employ projection-based DP-SGD that constraints noisy gradient updates to a learned $k$-dimensional subspace ($k \ll d$), reducing noise variance substantially. Second, we integrate Feature Differential Privacy (FDP), which relaxes differential privacy by decomposing gradient updates into public and private components, adding noise only to sensitive features. Finally, we propose a hybrid strategy that effectively combines projection and FDP, yielding multiplicative utility gains.

To summarise, our core contributions are as follows:
\begin{itemize}
    \item \textbf{First systematic DP benchmark for pose estimation:} We establish comprehensive baselines for differentially private 2D-HPE across privacy budgets ($\varepsilon \in \{0.2,0.4,0.6, 0.8\}$), clipping thresholds ($C \in \{0.01, 1.0\}$), and training strategies on MPII and HumanART datasets.
    \item \textbf{Feature-Projective Private Learning:} We propose a joint mechanism that integrates two complementary noise reduction strategies, First, a public dataset is used to identify a low dimensional gradient subspace to filter noise. Second, from Feature Differential Privacy (FDP), we define the entire raw image as private and add noise only to this while simultaneously using its corresponding public feature to compute a noise free gradient that helps with improved utility.
    \item \textbf{Convergence Analysis of Feature-Projective DP:} We show theoretically that the combined effect of projection and FDP is multiplicative in terms of signal-to-noise ratio and convergence speed. 
\end{itemize}

We conduct extensive experiments across diverse privacy budgets, clipping thresholds and training strategies on both MPII and HumanART datasets. The proposed feature-projective framework outperforms vanilla DP-SGD, demonstrating superior privacy-utility trade-offs. 

\section{Related Work}
\subsection{2D Human Pose Estimation and Privacy}
Markerless 2D-HPE  identifies anatomical keypoints in images without physical markers, playing a fundamental role in human motion analysis for healthcare and activity recognition ~\cite{deligianni2019emotions, bondugula2023novel}. While traditional heatmap methods based on CNN architectures \cite{artacho2020unipose, kamel2020hybrid, wang2022low, xiao2018simple} and vision-based transformers \cite{li2021test, li2021tokenpose, xu2022vitpose,purkrabek2025probpose} achieve state-of-the-art performance, they suffer from quantization errors and usually result in large cumbersome networks that are prone to memorization. Regression approaches offer faster, end-to-end solutions but with reduced accuracy \cite{toshev2014deeppose,  nie2019single, li2021human}.
Coordinate classification approach addresses some of these limitations by treating pose estimation as classification over discretized coordinates \cite{li2022simcc}, achieving strong performance with computational efficiency. Knowledge distillation techniques \cite{ye2023distilpose,li2021online,zhang2023human, kaushi2024BMVC} further enhance efficiency and inference speed by transfering knowledge from large models to compact architectures. 

Protecting user privacy remains critical yet challenging for HPE, which relies on high-quality images. Recent work demonstrates that adversaries can reconstruct substantial portions of private training data solely by analyzing the parameters or gradients of a trained neural network \cite{tian2025simudy,haim2022reconstructing, zhu2019deep}.
Consequently, data sharing in sensitive domains provide limited information, preventing analysis of crucial clinical indicators that require body shape or facial information for stress assessment and abnormal motion detection. 
Most platforms implement rudimentary privacy protection through face blurring and pixelation \cite{Ruiz-Zafra2023}, while more sophisticated methods include skin removal \cite{Hesse_2018_ECCV_Workshops} and template-based shape modeling, and visual privacy layers that degrade private attributes \cite{Hinojosa2021}. However, these ad hoc methods lack formal privacy guarantees and suffer from the "onion effect"~\cite{NEURIPS2022_OnionEffect}, where removing one protection layer exposes previously secured features, making them vulnerable to privacy attacks~\cite{zakariyya2025differentially}. 
Recent GAN-based anonymization techniques \cite{hukkelaas2023deepprivacy2, hukkelaas2023realistic,hukkelaas2023does} generate realistic anonymized figures, yet introduce artifacts due to pose detection errors and contextual mismatches. 
Adversarial learning approaches~\cite{xu2023userdp,maeng2023bounding} attempt to learn privatized features that obscure sensitive attributes while preserving task utility. However, these methods assume original data are unnecessary post-training, compromising interpretability, which is essential for clinical validation and diagnosis in healthcare applications. Moreover, even advanced anonymization cannot fully replace real data for training robust computer vision models~\cite{hukkelaas2023does}, highlighting the need for formal privacy frameworks like differential privacy that provide quantifiable guarantees without sacrificing data authenticity.
\subsection{Utility vs Privacy with DP optimization}
DP-SGD~\cite{abadi2016deep}  is one of the most common methods in developing privacy-preserved deep learning models because of the strong privacy guarantees compared to data-independent methods and its ability to scale to large datasets \cite{hardt2010geometry,dwork2021promise}. 
However, DP settings induce a fundamental privacy-utility trade-off that compromise practical deployment of privacy preserved HPE \cite{Abbasi2025}. DP-SGD performance depends on loss function smoothness \cite{pmlr-v107-wang20a}, gradient dimensionality \cite{fang2023improved,zhou2021bypassing} and clipping threshold selection \cite{lebensold2024privacy}. It has also been shown that if the loss function lacks Lipschitz continuity, the performance of DP-SGD critically depends on carefully selecting an appropriate clipping threshold; otherwise, performance will not improve significantly, regardless of the amount of training data or iterations \cite{fang2023improved}.
Recent research has sought to improve the utility-privacy trade-off by relaxing traditional differential privacy \cite{shi2021selective,golatkar2022mixed,mahloujifar2025machine}. \cite{shi2021selective} enhanced the utility of DP by introducing Selective Differential Privacy, which protects only sensitive tokens within language models, while leaving non-sensitive elements unperturbed. \cite{mahloujifar2025machine} proposed Feature Differential Privacy, a generalized DP framework that explicitly categorizes features as either protected or public, enabling targeted noise application that enhances utility. Both methods demonstrate that targeted protection of sensitive attributes substantially mitigates the privacy-utility trade-off compared to classical DP.
\begin{figure*}[ht]
\centering
\includegraphics[width=1.025\textwidth]{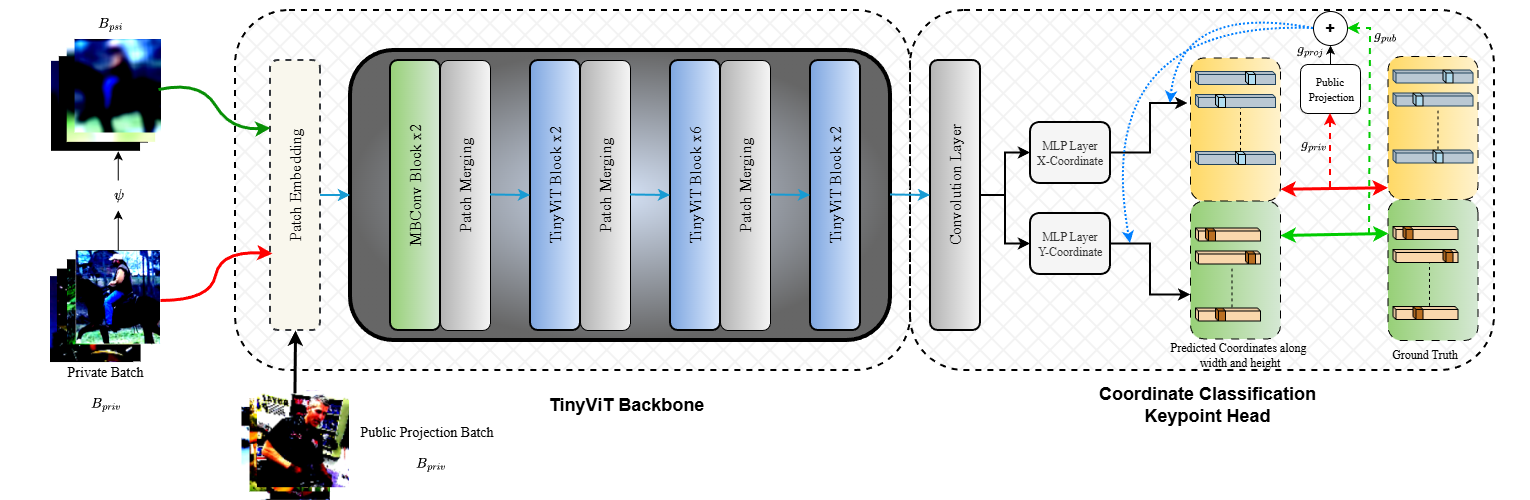}
\caption{Overview of our private HPE pipeline coupling a TinyViT based backbone with Coordinate Classification Keypoint head. The Public feature batch $B_{pub}$ is generated from the private batch $B_{priv}$ using $\psi$ both of which are given as input in a single iteration. Additionally, a public image set $B_{pub}^{proj}$ independent of $B_{priv}$ is used to calculate public gradients for projection at specified intervals. \textcolor{red}{Red Arrow} indicates the propagation of private gradients and \textcolor{green}{Green Arrow} indicates propagation of public feature gradients. \textcolor{cyan}{Blue Dotted Arrow} indicates the propagation of cumulative denoised gradient from Feature Projective DP.}
\label{fig:Arch}
\end{figure*}
\section{Methodology}
\subsection{Preliminaries: Differentially Private Stochastic Gradient Descent (DP-SGD)}
Privacy-preserving machine learning requires a rigorous mathematical framework to quantify privacy guarantees. DP provides such a framework by measuring in the context of databases how much the inclusion or exclusion of a single data point can influence the output of a randomized algorithm ~\cite{dwork2021promise,hardt2010geometry}. This comparison allows us to bound the information leakage about any individual data point. 

Consider two neighboring datasets-identical except for a single sample. The level of DP ensured by a randomized algorithm $\mathcal{M}$ is provided by the following definition.

\noindent\textbf{Definition 1}: $(\varepsilon, \delta)$-Differential Privacy:
A randomized algorithm $\mathcal{M}$ with domain $\mathcal{D}$ and range $\mathcal{R}$ is said to be $(\varepsilon, \delta)$-DP if, for any subset $S \subseteq \mathcal{R}$ and for any neighboring datasets $d, d' \in \mathcal{D}$, the following condition holds:
\begin{equation}
    \mathbb{P}[\mathcal{M}(d) \in S] \leq e^{\varepsilon} \cdot \mathbb{P}[\mathcal{M}(d') \in S] + \delta
\end{equation}
In this definition, $\varepsilon$ (privacy budget) controls the strength of the privacy guarantee. Smaller values of $\varepsilon$ provide stronger privacy. The parameter $\delta$ represents the probability that the privacy guarantee fails and is typically set to be cryptographically small.

In the context of HPE, differential privacy ensures that the inclusion or exclusion of a single training image, which potentially contains identifiable biometric information, does not significantly affect the model's learned parameters or predictions. Therefore, DP trained models provide a formal measure of privacy protection~\cite{dwork2021promise}, effectively mitigating risks from various attacks, including membership inference\cite{shokri2017membership} or reconstruction attacks\cite{zhu2019deep}. 

A common approach for ensuring differential privacy during neural network training is DP-SGD, which enforces an $(\varepsilon, \delta)$-DP guarantee on gradient updates~\cite{abadi2016deep, dupuy2022efficient, boenisch2024have}. This mechanism involves clipping gradients to a fixed $L2$-norm threshold ($C$) and adding Gaussian noise calibrated based on desired privacy budget ($\varepsilon$, $\delta$). This process ensures that no single training sample can disproportionately influence the model update~\cite{kong2023unified}. 

\subsection{Architecture}
For our 2D-HPE models, we adopt TinyViT \cite{Kan2022} as the backbone. This compact, four-stage efficient hierarchical vision transformer is well suited for resource-constrained vision tasks. Its smaller size is highly beneficial as the error bounds of DP-SGD are known to scale with number of parameters\cite{bassily2014private}. The model adopts a multi-stage architecture where in the spatial resolution is progressively reduced and the feature representation expands. TinyViT follows a hybrid architectural design containing convolutional layers at the initial stages followed by self-attention mechanisms. Unlike standard ViT models, TinyViT employs a two-layer convolutional embedding. In the first stage of the network, it employs MBConv \cite{Howard_2019_ICCV} blocks from MobileNetV2 to efficiently learn the low-level representation. The last three stages consists of transformer blocks hierarchically. Each stage consists of multi-head-self-attention (MHSA) layers, feed forward network (FFN) and $3 $ x $3$ depthwise convolutions between the MHSA and FFN layers.

For keypoint localization, we augment the TinyViT backbone model with a coordinate classification output stage~\cite{li2022simcc}. Given an input image \( I \in \mathbb{R}^{C \times H \times W} \) and a ground truth keypoint \( p_i = (x_i, y_i) \) for the \( i^\text{th} \) joint, the continuous coordinates are quantized into discrete bins via a splitting factor \( k \ge 1 \). Formally, the quantized coordinates are computed as:
\[
p'_i = \left( \left\lfloor x_i \cdot k \right\rfloor,\, \left\lfloor y_i \cdot k \right\rfloor \right),
\]
where \(\lfloor \cdot \rfloor\) denotes the rounding operation. This binning reduces quantization error while preserving high localization precision. The complete architecture is depicted in Figure \ref{fig:Arch}.

Within our network, the Convolutional head produces a 16-channel feature map, with each channel corresponding to a specific joint. These joint-specific features are Upsampled and flattened to form a compact representation used for classification over the discrete coordinate bins. To improve robustness, we employ Gaussian label smoothing on the classification targets. This smoothing accounts for spatial correlations by assigning soft labels that reflect the relevance of neighboring bins.
Finally, the discrete classification outputs are decoded back into continuous coordinates to yield the final keypoint predictions.

\subsection{Projection Based DP-SGD} \label{sec:proj}

Training dynamics in deep networks exhibit intrinsic low-dimensional structure, where meaningful gradient updates concentrate within a subspace significantly smaller than the full parameter space. We leverage this by identifying and projecting noisy gradients onto informative subspaces, filtering out less relevant directions while preserving signal quality under differential privacy constraints. In this way, we preserve the signal quality of gradient updates while adhering to DP constraints \cite{zhou2020bypassing}.
To estimate the intrinsic structure of the gradient space, we employ a small auxiliary public dataset $S_{pub}$, which is drawn from a similar distribution as that of private training set. This subset is used to estimate the principal subspace of the gradient covariance. Given the model parameters $w\in\mathbb{R}^{p}$, the second moment matrix of gradients over $S_{pub}$ is calculated as:
\begin{equation}
    M(w) = \frac{1}{m}\sum_{i=1}^{m}\nabla l(w,\tilde{z_i})\nabla l(w,\tilde{z_i})^{T}
\label{eq:secmom}
\end{equation}
where $m$ denotes the number of public samples and $\tilde{z}_i$ represents an input sample from $S_{pub}$. The eigenvectors corresponding to the top $k$ eigenvalues are stacked to form the projection matrix $\hat{V} \in \mathbb{R}^{p \times k}$ which forms the the low-dimensional approximation of the full gradient space; this maps the $p$-dimensional gradients to a smaller $k$-dimensional subspace. This projection matrix is updated periodically to accommodate changes in gradient distributions over the training period.

In the DP-SGD setup, for each mini-batch sampled from the private dataset $S_{priv}$, per-sample gradients are computed and the sensitivity of each individual gradient is bounded by the clipping threshold $C$:
\begin{equation}
    \tilde{g}_i = clip(\nabla l(w,z_i),C)
\end{equation}
The clipped gradients are aggregated over the batch and Gaussian noise is added to ensure differential privacy
\begin{equation}
    g = \frac{1}{B}  \left(\sum_{i \in B}\tilde{g}_i + \mathcal{N}(0, \sigma^2 C^2 \mathbf{I})\right), 
\end{equation}
where $B$ is the size of the mini-batch and $\sigma$ is the standard deviation. 
The full noisy gradient $g$ is then projected on to the estimated low-dimensional subspace as
\begin{equation}
    g_{proj} = (\hat{V}\hat{V}^T)g
\label{eq:proj}
\end{equation}

This restricts the update direction to the subspace where the gradients exhibit the highest variance, thereby filtering out noise components residing in less informative directions. The model parameters are then updated using the projected gradient. Since the projection is applied as a post-processing step after noise addition, the overall DP guarantee remains intact.

\subsection{Feature Projective DP-SGD for HPE} \label{sec:fdp}
\subsubsection{Feature Differential Privacy}
To enhance the model utility in our 2D HPE task, we extend the standard DP-SGD framework using Feature Differential Privacy (FDP). FDP exploits the transformation of the training image into private and public variants, selectively applying differential privacy only to sensitive features while freely utilizing non-sensitive(public) information \cite{mahloujifar2025machine}. 
Formally, let each sample be $x_i \in S_{data}$(a raw training image with keypoint labels), and let $\psi: S \rightarrow \mathcal{F}$ be a public feature map such that $\psi(x_i)$ is the public variant of the raw image $x_i$ and let $f: [0,1] \rightarrow[0,1]$ be a trade-off function. A randomized mechanism $\mathcal{M}$ satisfies $f$-FDP with respect to $\psi$ if, for any two datasets $d,d'$ differing in exactly one image-label pair $x_i \neq x_i'$ but having identical public representations $(\psi(x_i) = \psi(x_i'))$ and for all subsets $S$ for range of $\mathcal{M}$:
\begin{equation}
    \mathbb{P}[\mathcal{M}(d) \in S] \leq 1- f(\mathbb{P}[\mathcal{M}(d') \in S])
\end{equation}
Then, we say the mechanism is $(\varepsilon, \delta)$-DP with respect to $\psi$ iff it is $f$-FDP for $f(x) = 1- \delta -e^{\varepsilon}x$.
Motivated by this definition, the FDP-SGD method, explicitly distinguishes between public and private (raw) images to improve pose estimation accuracy under the same privacy budget as standard DP. Specifically, for each image-keypoint pair, we define a public loss $l_{pub}(w,\psi(x))$ which captures the coarse pose estimation based on the definition of $\psi$. The private loss $l_{priv}(w,x)$ captures the sensitive, fine-grained details of the human that requires privacy protection. Then the overall loss can be given as:
\begin{equation}
l(w,x) = l_{priv}(w,x) + l_{pub}(w,\psi(x))    
\label{eq:decomp}
\end{equation}

\subsubsection{Training with Feature-Projective DP}
To maximize the privacy utility tradeoff, we introduce Feature-Projective DP, a hybrid approach that integrates the two approaches outlined in Sections \ref{sec:proj}, \ref{sec:fdp}. The complete algorithmic details of this integrated approach are provided in Algorithm \ref{alg:pdp-sgd-mpii}. 

This approach synergizes two key ideas: First we adopt the FDP framework to decompose the total loss into a public component $l_{pub}$ (computed on public features $\psi(x)$) and a private component $l_{priv}$ (computed on raw image $x$). This ensures that DP noise is added only to the gradient of sensitive private component. Second, we apply the projection technique to filter noise, restricting the private gradient update to the most informative $k$-dimensional subspace. As shown in Algorithm \ref{alg:pdp-sgd-mpii}, our Feature-projective DP method proceeds at each iteration $t$ by first sampling two separate and independent batches from $S_{data}$.  On the public batch $B_{psi}^t$ we compute the gradient as:
\begin{equation}
    g_{pub}^t = \frac{1}{|B_{psi}^t|} \sum_{x\in B_{psi}^t} \nabla l_{pub}(w_{t-1},\psi(x))
\end{equation}
Similarly, on the private batch $B_{priv}^t$, we compute and clip the gradient of the private loss to the clipping norm $C$ as $\tilde{g}$, then aggregate and add gaussian noise:
\begin{equation}
    g_{priv}^t = \frac{1}{|B_{priv}^t|} \left(\sum_{x\in B_{priv}^t} \tilde{g} + \mathcal{N}(0,\sigma^2C^2I)\right)
\end{equation}
We then denoise $g_{priv}^t$ by applying the subspace projection from Eq.\ref{eq:proj} as a post-processing step given as:
\begin{equation}
    g_{proj}^t = (\hat{V}_t\hat{V}_t^T)g_{priv}^t
\end{equation}
The final gradient update $g_t$ is the sum of the clean public component and the denoised private component. The model parameters are then updated as:
\begin{equation}
    g_t = g_{pub}^t + g_{proj}^t 
\end{equation}
\begin{equation}
    w_t = w_{t-1} - \eta_t g_t
\end{equation} where $\eta_t$ denotes the learning rate.

\subsubsection{Convergence Analysis of Feature-Projective DP}
The convergence analysis of our method formally establishes the utility gain as observed from our empirical results and is a direct corollary of the separate analyses from \cite{zhou2020bypassing,mahloujifar2025machine}.

Let the empirical risk be $\hat{L}_n(w) = \frac{1}{n}\sum_{i=1}^{n}l(w,x_i)$ on a private dataset $S_{priv}$ of size $n$. 

\noindent\textbf{Assumption 1.} The loss $l(w,x)$ can be decomposed into public and private components as given in Eq.~\ref{eq:decomp}.

\noindent\textbf{Assumption 2.} The full loss $\hat{L}_n(w)$ is $\rho$-smooth, the full gradient $||\nabla l(w,x)||_2 \leq G$ is bounded where $G$ defines the sensitivity for subspace reconstruction error and the private gradient is bounded by the threshold $C$ as $||\nabla l_{priv}(w,x)||_2 \leq C$, where $C \leq G$.

\noindent\textbf{Assumption 3.} We have access to a separate public dataset $S_{pub}$ of size $m$ and $\hat{V}_t  \in \mathbb{R}^{p \times k}$ is the $k$-dimensional projection matrix computed from top-$k$ eigenspace (from Eq.~\ref{eq:secmom}) on $S_{pub}$ at iteration $w_{t-1}$.

\noindent\textbf{Assumption 4.} Assuming the principal component of the gradient dominance condition is satisfied and under this, we denote the eigengap at iteration $t$ as $\alpha_t$ and $\Lambda = \frac{1}{T}\sum_{t=1}^{T}1/\alpha_t^2$ be average inverse squared eigengap and refer to $\gamma_2(\mathcal{W}, d_w)$ as the associated complexity measure (where $\mathcal{W}$ iterate set of the weights and $d_w$ is distance between them), as defined in \cite{zhou2021bypassing}. 

Under these assumptions, setting the total iterations $T = \mathcal{O}(n^2\varepsilon^2)$, the average expected gradient norm of feature-projective DP is bounded by:
{\scriptsize
\begin{equation}
    \frac{1}{T}\sum_{t=1}^{T}\mathbb{E}||\nabla\hat{L}_{n}(w_{t})||_{2}^{2} \le \underbrace{\tilde{\mathcal{O}}\left(\frac{k \cdot \rho \cdot C^2}{n\varepsilon}\right)}_{\text{Privacy Error}} + \underbrace{\mathcal{O}\left(\frac{\Lambda G^{4}\rho^{2}\gamma_{2}^{2}(\mathcal{W},d_w)\ln p}{m}\right)}_{\text{Reconstruction Error}}
\end{equation}
}

The convergence is bound by two terms: a reconstruction error inherited from use of public dataset $S_{pub}$ and privacy error from the gaussian noise. By combining both the approaches, the privacy error scales with both the reduced dimension $k$ and reduced gradient norm $C$ which can be understood from the error bound changing from $\tilde{\mathcal{O}}(p \cdot G^2) \rightarrow \tilde{\mathcal{O}}(k \cdot C^2)$ which explains the feature-projective DP's higher utility for the same $(\varepsilon, \delta)$-FDP guarantee.

\begin{algorithm}
\caption{Feature Projective DP-SGD}
\label{alg:pdp-sgd-mpii}
\begin{algorithmic}[1]
\REQUIRE Full dataset $\mathcal{S}_{\rm data} = \{x_1, \dots, x_n\}$, split into 
public subset $\mathcal{S}_{\rm pub}\subset\mathcal{S}_{\rm data}$ (size $m$) and 
private remainder $\mathcal{S}_{\rm priv}=\mathcal{S}_{\rm data}\setminus\mathcal{S}_{\rm pub}$, public feature map $\psi$,\\
combined loss $\ell(w,z)$ with public and private losses $l_{pub},l_{priv}$, clip norm $C$, noise std.\ $\sigma$, subspace dim $k$,\\
batch size $B$, iterationss $T$, learning rate $\{\eta_t\}$.

  \STATE Initialize model parameters $w_0\in\mathbb R^p$.
  \FOR{$t=1,\dots,T$}
    \vspace{0.5mm}\hrule\vspace{0.5mm}
  \STATE \textbf{(1) Subspace identification on $\mathcal S_{\rm pub}$:}
  \STATE Compute 
    $$
      M_t \;=\;\frac1m\sum_{z\in\mathcal S_{\rm pub}}
        \nabla\ell(w_{t-1},\tilde{z})
        \,\nabla\ell(w_{t-1},\tilde{z})^\top.
    $$
  \STATE Compute the top-$k$ eigenvectors of $M_t$.
  \STATE Form the subspace basis 
    $\hat{V_t}\in\mathbb R^{p\times k}$.
  \STATE Compute the projector 
    $\hat{V_t}\,\hat{V_t}^\top\in\mathbb R^{p\times p}$.
  \hrule\vspace{0.5mm}
  \STATE \textbf{(2) Compute public and private feature gradient:}
    \STATE Sample public batch $B_{psi}^{t} \subset {\psi(x):x \in S_{priv}}$
    \STATE Compute 
    $$
    g_{pub}^{t} = \frac{1}{|B_{psi}^t|} \sum_{x\in B_{psi}^t} \nabla l_{pub}(w_{t-1},\psi(x))
    $$
    
    \STATE Sample private batch $B_{priv}^t \subset S_{priv}$
    \STATE Compute the clipped gradient $\tilde{g_t}$, aggregate and add Gaussian noise
    $$
    g_{priv}^t = \frac{1}{|B_{priv}^t|} \left(\sum_{x\in B_{priv}^t} \tilde{g_t} + \mathcal{N}(0,\sigma^2C^2I)\right)
    $$
    \STATE \emph{Project:}\quad 
      $g_{proj}^t = (\hat{V_t}\,\hat{V_t}^\top)\cdot\,g_{priv}^t\in\mathbb R^p$.
    \STATE Merge Public and Private projected feature gradients
    $$
    g_t = g_{pub}^t + g_{proj}^t
    $$
    \STATE \emph{Update:}\quad 
      $w_t = w_{t-1} - \eta_t\,g_t$.
  \ENDFOR
  \STATE \textbf{return} $w_T$.
\end{algorithmic}
\end{algorithm}


\section{Experiments}
\subsection{Dataset and Implementation Details}
In our experiments, we evaluated our framework on two widely used human pose datasets: MS COCO Keypoint Dataset \cite{lin2014microsoft} and MPII dataset \cite{andriluka20142d}. Our methodology assumes that the COCO dataset serves as a public dataset used for pre-training the network weights, while MPII/ HumanART functions as a private dataset on which we apply the differential privacy techniques. 

Specifically, our models are pretrained on the COCO \textit{train2017} set, which consists of approximately 118k images with around 140k annotated human instances, each with 17 joint annotations. The \textit{val2017} set consisting of around 5k images is used for validation. For evaluating the trade-off between utility and performance under various DP-SGD techniques we employ the MPII Human Pose Dataset consisting of 40k human instances, each labeled with 16 joint annotations. When transferring the model from COCO to MPII, we adjust for the keypoint discrepancy between datasets. We employ the Percentage of Correct Keypoints normalized by head (PCKh) \cite{andriluka20142d} as an evaluation metric. 

To further assess the generalization of our privacy-preserved pose estimation models under domain shift and visual diversity, we conduct additional experiments on the Human-Art dataset \cite{ju2023human}. Human-Art is a recently introduced, large-scale human-centric benchmark designed to bridge natural and artificial visual domains. It contains 50,000 high-quality images with over 123,000 person instances across 20 diverse scenarios, spanning natural scenes (e.g., cosplay, drama, dance) and a wide spectrum of artistic styles (e.g., oil paintings, sculptures, digital art, watercolor, and murals). 
Compared to conventional datasets like MPII or COCO, Human-Art presents significantly greater challenges for pose estimation due to the presence of stylized or abstract human depictions, exaggerated or distorted body proportions, occlusions, artistic textures, and unconventional poses. We follow the standard MS COCO evaluation protocol and report the Average Precision (AP) as the primary metric.

Our experimental framework explores three distinct DP training scenarios: Fine-Tuning with frozen backbone, Full Fine-Tuning and, Training from scratch. For the first scenario, we specifically freeze the first three stages of the backbone and finetune the fourth stage and all instances of layer norm\cite{de2022unlocking}. To generate the public feature map, we employ Gaussian blur as $\psi$ which effectively suppresses facial and body structure details. Details on datasets, training and privacy related parameters are provided in the supplementary.

\begin{figure}[!htb]
\centering

\includegraphics[width=0.48\textwidth]{CVPR_figure_full_with_public_features.png}
\caption{Comparison of PCKh@0.5 on MPII dataset across private and non-private methodologies under different training strategies with varied privacy budget ($\varepsilon$) and clipping thresholds ($C$).\footnotemark}
\label{fig:results}
\end{figure}

\begin{figure}[!htb]
\centering
\includegraphics[width=0.40\textwidth]{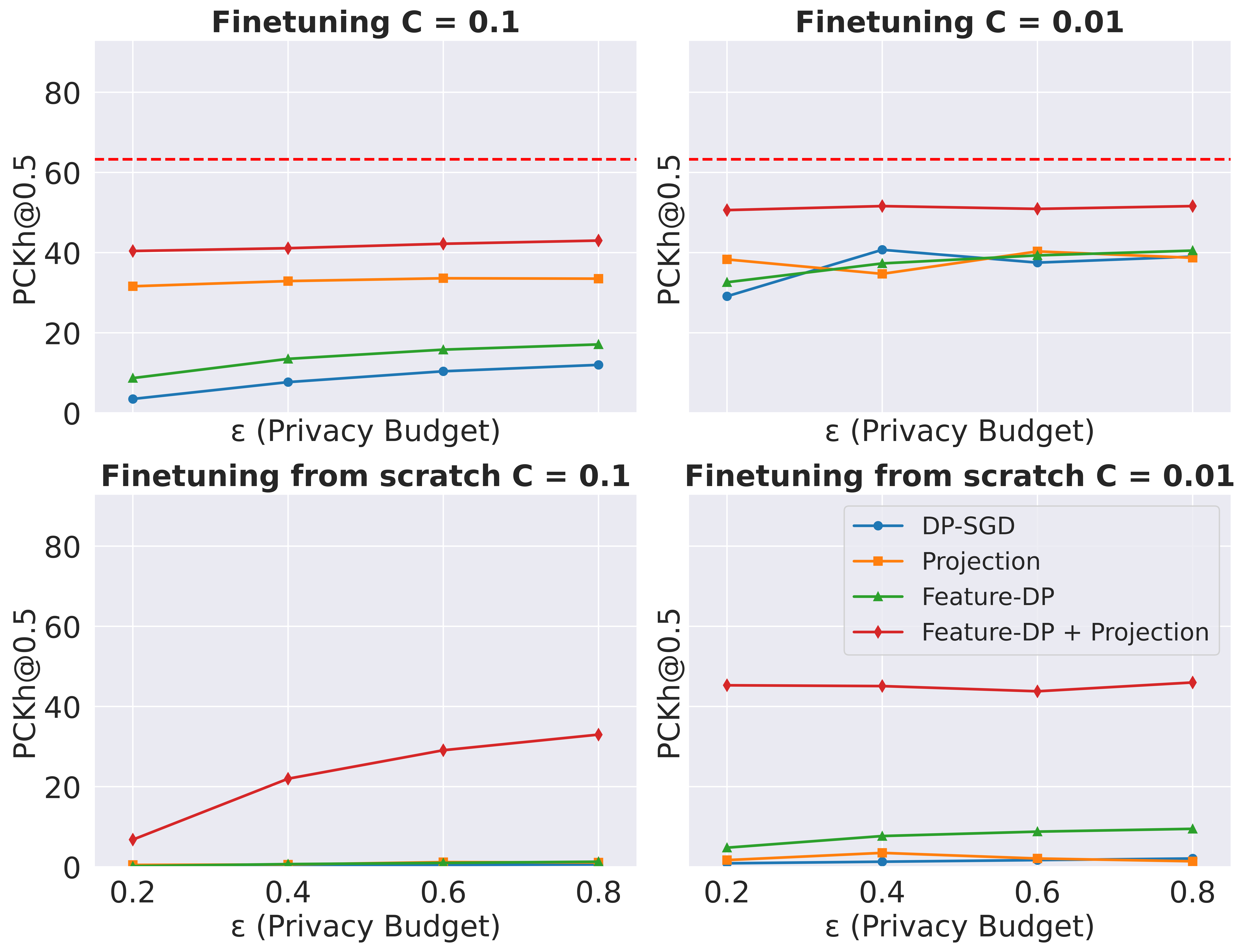}
\caption{Comparison of AP on HumanART dataset across private and non-private methodologies under different training strategies with varied privacy budget ($\varepsilon$) and clipping thresholds ($C$).\footnotemark[\value{footnote}]}
\label{fig:results_HART}
\end{figure}
\footnotetext{Full tabular results are provided in the supplementary material.}

\subsection{Results and Analysis on MPII dataset}
For comprehensive comparison, all experimental results across training strategies, clipping thresholds and privacy budgets are visualized in Figure \ref{fig:results}.

\subsubsection{\textbf{Non-Private Baseline Results}}
Table \ref{tab:mpii_non_dp_results} presents baseline pose estimation performance of our model on the MPII dataset under three training strategies: (i) finetuning from a COCO-pretrained model, (ii) finetuning from scratch (initialization with COCO pretrained weights and all layers are trained), and (iii) training from scratch (random initialization). Additionally, we report results from using only public features (blurred images) under the same strategies to provide context for evaluating privacy-utility trade-offs.
\begin{table}[h]
\centering
\caption{MPII Results: Non-Private Baselines for our HPE model on the MPII dataset}
\label{tab:mpii_non_dp_results}
\large
\footnotesize
\setlength{\tabcolsep}{3pt}
\renewcommand{\arraystretch}{1.3}
\resizebox{\columnwidth}{!}{%
\begin{tabular}{l |c| c| c| c| c| c| c| c| c}
\toprule
Training Strategy & Head & Shoulder & Elbow & Wrist & Hip & Knee & Ankle & Mean & Mean@0.1 \\
\midrule
\textbf{Finetuning} & 97.07 & 95.86 & 89.59 & 83.61 & 89.29 & 85.31 & 81.48 & 89.36 & 31.33 \\
\textbf{Finetuning from scratch} & 96.45 & 95.84 & 88.07 & 82.18 & 88.78 & 83.01 & 79.45 & 88.28 & 28.11 \\
\textbf{Training from scratch} & 93.89 & 89.32 & 75.34 & 65.24 & 80.04 & 66.15 & 60.60 & 76.89 & 17.26 \\
\textbf{Finetuning on Public features} &94.30&93.95&83.21&75.19&83.53&76.86&72.76&83.61&20.81 \\
\textbf{Finetuning from scratch on Public features}&88.71&84.32&69.71&54.37&70.24&61.67&55.12&70.32&11.36\\
\textbf{Training from scratch on Public features}&15.31&20.01&17.19&12.59&25.04&16.90&10.91&17.99&0.78\\
\bottomrule
\end{tabular}
}
\end{table}
As expected, the finetuning strategy achieves the highest mean accuracy of 89.36\% followed by finetuning from scratch (88.28\%) and training from scratch (76.89\%), which is to be expected. These non-private baselines establish upper bound performance references for evaluating differential privacy impact. When the model relies only on public features(gaussian blurred images), performance reduces significantly. While finetuning on public features maintain reasonable accuracy, the other training strategies yield substantially compromised results, confirming that fine-grained visual details in raw images are critical for accurate pose estimation.

\begin{figure*}[!htb]
\centering
\includegraphics[width=1.0\textwidth]{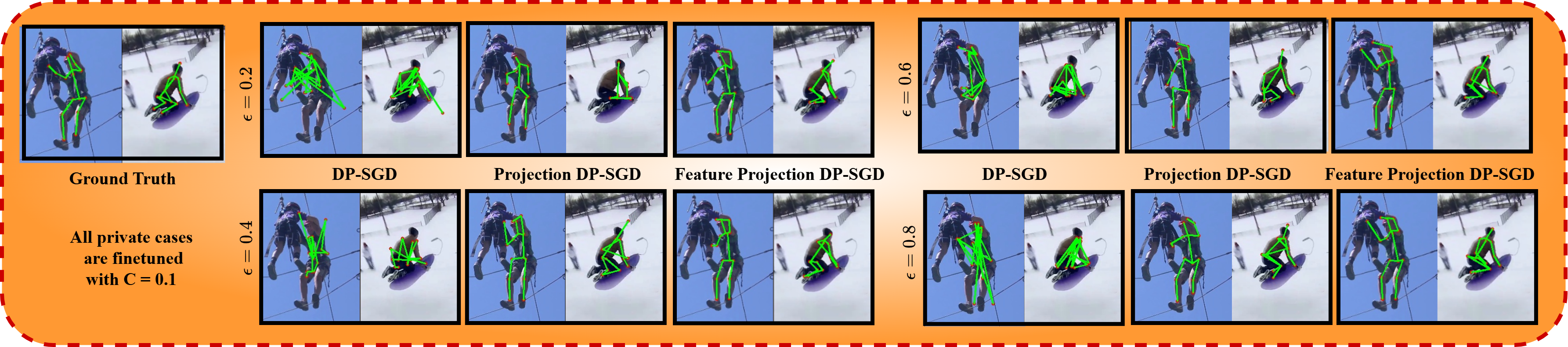}
\caption{Depiction of qualitative results on DP-SGD, Projection DP-SGD and Feature Projection DP-SGD. We specifically show results on Finetuning with $C=0.1$ at various privacy budgets.}
\label{fig:qual}
\end{figure*}

\subsubsection{\textbf {DP-SGD Baseline Results}}
Figure \ref{fig:results} presents the PCKh@0.5 results on MPII dataset under the aforementioned training strategies using DP-SGD. Experiments were conducted across multiple settings with varying privacy parameters ($\varepsilon \in \{0.2,0.4,0.6,0.8\}$) and clipping thresholds ($C \in \{0.01,0.1,1.0\}$).
For standard finetuning with DP-SGD, lower clipping thresholds consistently yield better pose estimation results across different privacy levels. Specifically, at $C = 0.01$, the model achieves substantially higher accuracy of 63.85\% mean PCKh@0.5 at the tightest privacy loss($\varepsilon = 0.2$) compared to $C = 0.1$ (28.46\%) and $C=1.0$ (5.94\%). This is indeed because of the fact that the effective noise magnitude grows linearly with the $C$ thus our results confirm this.

Notably, finetuning the COCO-pretrained TinyViT backbone significantly mitigates the DP induced performance degradation compared to training from scratch or finetuning from scratch \cite{yu2021differentially}. This indicates that pretrained human pose based feature representations provide robust feature priors that enable DP-SGD to adapt effectively to private pose datasets, while maintaining resilience to noise corruption.

\subsubsection{Performance Analysis of Subspace Projection}
We maintain identical training strategies and privacy parameters to ensure direct comparison with both non-private and DP-SGD baseline methods. 
Our subspace projection approach demonstrates substantial performance improvements across multiple configurations.
At the most restrictive clipping threshold ($C=0.01$), projection yields significant gains from 63.85\% to 78.48\% at $\varepsilon=0.2$ and from 78.17\% to 80.63\% at $\varepsilon = 0.8$. 
This enhancement occurs because, while Gaussian noise is injected uniformly across all gradient components, only a subset of directions carry meaningful pose-relevant information. By projecting onto the learned subspace, we effectively discard the noise in irrelevant directions, thereby improving signal-to-noise ratio and preserving essential pose estimation features.
At $C = 0.1$, the projection approach consistently outperforms baseline DP-SGD. Finetuning increases accuracy from 73.13\% to 77.41\%, while training from scratch improves from 9.80\% to 13.05\%. However, for the finetuning from scratch strategy, the curve plateaus slightly below regular DP-SGD. We attribute this phenomenon to the interaction between injected gaussian noise and subspace reconstruction error\cite{zhou2020bypassing}. 
At the largest clipping threshold ($C=1.0$), we observe non-monotonic patterns. Under this condition, the raw gradients become dominated by noise, leading to unstable parameter updates and local dips in accuracy.

\subsubsection{Performance analysis of FDP and Feature-Projective DP}
Feature DP consistently outperforms vanilla DP-SGD across all experimental configurations. 
Under finetuning with $C = 0.01$, FDP achieves substantial improvements, from 63.85\% to 75.46\% at $\varepsilon = 0.2$ (11.61\% gain) and from 78.17\% to 80.40\% at $\varepsilon = 0.8$ (2.23\% gain). 
This is consistently observed across all training strategies and clipping values. 
Integrating FDP with subspace projection results in the highest accuracy across all experimental settings. 
Even under the most challenging conditions with stringent clipping of $C=1.0$, where standard DP-SGD achieves only 12.53\%, Feature-projective DP attains 71.66\%, representing a six fold relative gain.

The largest improvements occur when training from scratch. With $C=0.1$ and $\varepsilon = 0.8$, vanilla DP-SGD achieves merely 6.85\% accuracy, while FDP alone attains 11.22\%. However, the combined Feature-Projective DP approach achieves 33.48\%. This demonstrates that combining both techniques boosts utility drastically especially in large noise induced scenarios, where neither alone suffices to recover strong pose features from corrupted gradients. Figure \ref{fig:qual} depicts few qualitative results across different privacy strategies along with ground truth.

\subsection{Cross-Dataset Evaluation on HumanART}
We further evaluate our feature-projective DP framework on the HumanART dataset, which contains stylized and artistic human figures with substantial visual domain shifts relative to natural images in MPII. As shown in Figure \ref{fig:results_HART}, our method maintains a strong privacy-utility balance across all privacy budgets, achieving 51.6 mAP at $\varepsilon = 0.8$ with finetuning strategy at $C=0.01$. For clarity, we report only finetuning and finetuning from scratch at $C=\{0.01,0.1\}$, as these are the only settings that yield stable and practically useful performance. Training from scratch or using using $C=1.0$ yields negligible accuracy. Results are visualised in Figure \ref{fig:results_HART} while full tabular results are provided in supplementary. Notably, under non-private training, finetuning from scratch achieves higher accuracy (69.5 mAP) than finetuning by freezing the backbone (63.3 mAP), as expected from the greater capacity for task-specific adaptation. However, this trend reverses once DP is applied. We attribute this behavior to the addition of DP noise and clipping where updating a smaller subset of parameters concentrates the effective learning while reducing total injected noise. This observation aligns with prior findings\cite{shen2021towardss} that DP noise disproportionately harms larger parameter regimes.


\section{Conclusion}
Our work presents the first differentially private (DP) approach to 2D human pose estimation (HPE), addressing critical privacy concerns while maintaining utility. 
Our results clearly establish that the synergistic combination of feature-level privacy and subspace projection dramatically enhances utility across all settings.
Importantly, our proposed Feature-Projective DP 2D-HPE approach achieved up to 82.62\% mean PCKh@0.5 on MPII and 51.6 mAP on HumanART  at $\epsilon = 0.8$, significantly narrowing the gap to non-private performance under strong formal privacy guarantees.
Crucially, the proposed approach requires no manual curation of private features, as it automatically protects the entire raw image, ensuring privacy preservation for both individuals and their spatial environments.

\section{Acknowledgments}

We acknowledge funding from EPSRC(EP\verb|\|W01212X \verb|\|1), the Royal Society (RGS\verb|\|R2\verb|\|212199) and Academy of Medical Sciences (NGR1\verb|\|1678).
{
    \small
    \bibliographystyle{ieeenat_fullname}
    \bibliography{main}
}

 \appendix
  \begin{table}[h]
\centering
\resizebox{\columnwidth}{!}{%

\begin{tabular}{cl}
\toprule
\textbf{Symbol} & \textbf{Definition} \\
\midrule
\multicolumn{2}{l}{\textit{Data and Features}} \\
$\mathcal{S}_{\rm data}$ & Full dataset of size $n$ \\
$\mathcal{S}_{\rm pub}$ & Public subset for subspace estimation, size $m$ \\
$\mathcal{S}_{\rm priv}$ & Private subset for training, size $n-m$ \\
$x$ & Data sample with private features \\
$\psi(x)$ & Public feature map (non-sensitive transformation) \\
$B_{priv}$ & Private mini-batch \\
$B_{psi}$ & Public mini-batch for Feature-DP \\
\midrule
\multicolumn{2}{l}{\textit{Model and Loss}} \\
$w \in \mathbb{R}^p$ &  Total model parameters\\
$\ell(w,x)$ & Total loss function \\
$\ell_{\rm priv}(w,x)$ & Private loss component (depends on private features) \\
$\ell_{\rm pub}(w,\psi(x))$ & Public loss component (depends only on $\psi(x)$) \\
$\hat{L}_n(w)$ & Empirical risk: $\frac{1}{n}\sum_{i=1}^n \ell(w,x_i)$ \\
$\rho$ & Lipschitz constant of the gradient (smoothness parameter of loss) \\
\midrule
\multicolumn{2}{l}{\textit{Gradients and Sensitivity}} \\
$\nabla \ell(w,x)$ & Full gradient \\
$\nabla \ell_{\rm priv}(w,x)$ & Private gradient component \\
$\nabla \ell_{\rm pub}(w,\psi(x))$ & Public gradient component \\
$G$ & Full gradient sensitivity: $\|\nabla \ell(w,x)\|_2 \leq G$ \\
$C$ & Private gradient sensitivity: $\|\nabla l_{\rm priv}(w,x)\|_2 \leq C$ \\
$g^t$ & Gradient estimate at iteration $t$ \\
$\tilde{g}$ & Clipped gradient \\
\midrule
\multicolumn{2}{l}{\textit{Subspace and Projection}} \\
$k$ & Subspace dimension ($k \ll p$) \\
$M_t$ & Gradient covariance matrix at iteration $t$ \\
$\hat{V}_t \in \mathbb{R}^{p \times k}$ & Top-$k$ eigenvectors of $M_t$ (subspace basis) \\
$\lambda_i(M_t)$ & $i$-th eigenvalue of $M_t$ (ordered: $\lambda_1 \geq \lambda_2 \geq \ldots$) \\
$\alpha_t$ & Eigengap: $\lambda_k(M_t) - \lambda_{k+1}(M_t)$ \\
$\Lambda$ & Average inverse squared eigengap: $\frac{1}{T}\sum_{t=1}^T 1/\alpha_t^2$ \\
$\gamma_2(\mathcal{W}, d_w)$ & Complexity measure for set of all possible model iterates $\mathcal{W}$ \\
\midrule
\multicolumn{2}{l}{\textit{Privacy and Optimization}} \\
$(\varepsilon, \delta)$ & Differential privacy parameters \\
$\sigma$ & Gaussian noise standard deviation \\
$T$ & Total number of iterations \\
$\eta_t$ & Learning rate at iteration $t$ \\
\bottomrule
\end{tabular}
}
\caption{Summary of Notation}
\label{tab:notation}
\end{table}

\section{Implementation Details}
 Our models are pretrained on the COCO \textit{train2017} set, which consists of approximately 118k images with around 140k annotated human instances, each with 17 joint annotations. The \textit{val2017} set consisting of around 5k images is used for validation. For evaluating the trade-off between utility and performance under various DP-SGD techniques we employ the MPII Human Pose Dataset consisting of 40k human instances, each labeled with 16 joint annotations. When transferring the model from COCO to MPII, we adjust for the keypoint discrepancy between datasets. We employ the Percentage of Correct Keypoints normalized by head (PCKh) \cite{andriluka20142d} as an evaluation metric.

 To assess model generalization under significant domain shifts, we utilize the Human-Art dataset \cite{ju2023human}, a large-scale benchmark designed to bridge natural and artificial visual domains . The dataset comprises 50,000 images with over 123,000 instances across 20 diverse scenarios, encompassing both natural scenes (e.g., dance, drama) and artistic representations (e.g., oil paintings, digital art) . Human-Art presents unique challenges absent in conventional datasets like MPII, including abstract depictions, distorted body proportions, and unconventional poses . We adhere to the standard MS COCO evaluation protocol, reporting Average Precision (AP) as the primary metric .
 
 All models are trained under differential privacy constraints using DP-SGD with various clipping norms and privacy budgets ($\epsilon$). Each model undergoes training for a total of 25 epochs, as we empirically observed no significant performance improvements when extending training beyond this duration under DP constraints. Throughout all experiments, we maintain a fixed input resolution of $256 \times192$ pixels to ensure consistency across experiments and enable comparison with prior work.

 For the privacy parameter settings, we use three gradient clipping norms $C = \{1.0, 0.1, 0.01\}$, with target privacy budgets of $\epsilon = \{0.2,0.4,0.6,0.8\}$. We adopt Renyi Differential Privacy (RDP) \cite{mironov2017renyi} for privacy accounting with the privacy parameter $\delta = 4e-5$.

 For the projection method, we randomly select 100 samples from the training dataset of MPII as $S_{pub}$ (ensuring no image overlap with the private data) with the remaining data forming the private training set $S_{priv}$. The default projection dimension $K$ is set to 50 for all experiments. 

 To generate the public feature map, we employ Gaussian blur as $\psi$ with a kernel size of $(25,25)$ and standard deviation $\sigma_X = 10$, which effectively suppresses facial and body structure details. We deliberately blur the entire image rather than selectively masking human regions, as this approach provides comprehensive privacy protection by obscuring not only human identities but also contextual environmental details as depicted in Figure \ref{fig:qual}(e).

\subsection{Convergence Analysis of Feature-Projective DP}
To contextualize our contribution, we first present the privacy error bounds for the baseline methods. The convergence of standard DP-SGD for non-convex objectives is limited by a privacy error term that scales with the ambient dimension $p$ and the full gradient sensitivity $G$, specifically $\tilde{\mathcal{O}}(\frac{p \cdot G^2}{n\varepsilon})$ \cite{bassily2014private, wang2019differentially}. The two frameworks we synthesize target distinct factors of the privacy error:

\noindent \textbf{Gradient Projection (PDP-SGD)} \cite{zhou2020bypassing}: reduces the \textit{dimensional} dependence by projecting the noise onto a lower-dimensional subspace, replacing $p$ with the subspace dimension $k$. This yields a privacy error of $\tilde{\mathcal{O}}(\frac{k \cdot G^2}{n\varepsilon})$.

\noindent \textbf{Feature-DP (FDP)} \cite{mahloujifar2025machine}: reduces the \textit{magnitude} dependence by privatizing only the sensitive private loss component, replacing the full sensitivity $G$ with the private sensitivity $C$. This yields a privacy error of $\tilde{\mathcal{O}}(\frac{p \cdot C^2}{n\varepsilon})$.

The convergence analysis of our method formally establishes the utility gain as observed from our empirical results and is a direct corollary of the separate analyses from \cite{zhou2020bypassing,mahloujifar2025machine}.

Let the empirical risk be $\hat{L}_n(w) = \frac{1}{n}\sum_{i=1}^{n}l(w,x_i)$ on a private dataset $S_{priv}$ of size $n$. 

\noindent\textbf{Assumption 1.} The loss $l(w,x)$ can be decomposed into public and private components given as $\ell(w,x) = \ell_{priv}(w,x) + \ell_{pub}(w,\psi(x))  $.

\noindent\textbf{Assumption 2.} The full loss $\hat{L}_n(w)$ is $\rho$-smooth, the full gradient $||\nabla \ell(w,x)||_2 \leq G$ is bounded where $G$ defines the sensitivity for subspace reconstruction error and the private gradient is bounded by the threshold $C$ as $||\nabla \ell_{priv}(w,x)||_2 \leq C$, where $C \leq G$.

\noindent\textbf{Assumption 3.} We have access to a separate public dataset $S_{pub}$ of size $m$ and $\hat{V}_t  \in \mathbb{R}^{p \times k}$ is the $k$-dimensional projection matrix computed from top-$k$ eigenspace as $M(w) = \frac{1}{m}\sum_{i=1}^{m}\nabla l(w,\tilde{z_i})\nabla l(w,\tilde{z_i})^{T}$ on $S_{pub}$ (Eq.2 of main paper) at iteration $w_{t-1}$.

\noindent\textbf{Assumption 4.} Assuming the principal component of the gradient dominance condition is satisfied and under this, we denote the eigengap at iteration $t$ as $\alpha_t$ and $\Lambda = \frac{1}{T}\sum_{t=1}^{T}1/\alpha_t^2$ be average inverse squared eigengap and refer to $\gamma_2(\mathcal{W}, d_w)$ as the associated complexity measure (where $\mathcal{W}$ iterate set of the weights and $d_w$ is distance between them), as defined in \cite{zhou2021bypassing}. 

Under these assumptions, setting the total iterations $T = \mathcal{O}(n^2\varepsilon^2)$, the average expected gradient norm of feature-projective DP is bounded by:
{\scriptsize
\begin{equation}
    \frac{1}{T}\sum_{t=1}^{T}\mathbb{E}||\nabla\hat{L}_{n}(w_{t})||_{2}^{2} \le \underbrace{\tilde{\mathcal{O}}\left(\frac{k \cdot \rho \cdot C^2}{n\varepsilon}\right)}_{\text{Privacy Error}} + \underbrace{\mathcal{O}\left(\frac{\Lambda G^{4}\rho^{2}\gamma_{2}^{2}(\mathcal{W},d_w)\ln p}{m}\right)}_{\text{Reconstruction Error}}
\end{equation}
}

The convergence is bound by two terms: a reconstruction error inherited from use of public dataset $S_{pub}$ and privacy error from the gaussian noise. By combining both the approaches, the privacy error scales with both the reduced dimension $k$ and reduced gradient norm $C$ which can be understood from the error bound changing from $\tilde{\mathcal{O}}(p \cdot G^2) \rightarrow \tilde{\mathcal{O}}(k \cdot C^2)$ which explains the feature-projective DP's higher utility for the same $(\varepsilon, \delta)$-FDP guarantee.

\section{Per Joint PCKh@0.5 on MPII and AP on HumanART}

We provide an extensive evaluation of our proposed training strategies on the MPII dataset (Tables \ref{tab:mpii_results} - \ref{tab:mpii_blur_projection}), highlighting the impact of gradient clipping norm $C$, privacy budget $\varepsilon$, and initialization strategies across four setups: vanilla DP-SGD, DP-SGD with projection, Feature DP-SGD, and our proposed Feature Projective DP. In the baseline DP-SGD (Table \ref{tab:mpii_results}), fine-tuning from a pretrained model with \(C = 0.01\) and \(\epsilon = 0.8\) achieves a mean PCKh of 78.17\%, while the same configuration trained from scratch drops to 12.74\%, emphasizing the importance of pretrained representations under privacy constraints. Incorporating gradient projection (Table \ref{tab:mpii_new_results}) significantly improves results across all settings, with fine-tuning at \(C = 0.01, \epsilon = 0.8\) reaching 80.63\% mean PCKh, and even training from scratch improving to 13.96\%. Feature DP (Table \ref{tab:mpii_blur}), which perturbs only the private gradient component, also shows substantial gains over vanilla DP-SGD, particularly for pretrained models (80.41\% at \(C = 0.01, \epsilon = 0.8\)). However, the most consistent and robust performance is achieved with our Feature Projective DP method (Table \ref{tab:mpii_blur_projection}), which combines both projection and selective privatization. It reaches a peak mean PCKh of 82.50\% when fine-tuning with \(C = 0.01\) and \(\epsilon = 0.2\), and maintains performance above 81\% across all \(\epsilon\) values, which shows strong utility even under tighter privacy budgets. Notably, feature projective DP is the only setup where fine-tuning from scratch reaches competitive accuracy (79.95\% at \(\epsilon = 0.8\)), significantly outperforming all baselines and confirming its ability to generalize under both high privacy constraints and limited initialization. Additionally, we provide the qualitative results in Figure \ref{fig:qual}. These results collectively validate that our combined approach effectively mitigates utility loss inherent in private learning for structured prediction tasks such as human pose estimation.

On the HumanART dataset, we evaluate our differentially private methods under the same four configurations as MPII: DP-SGD, DP-SGD with projection, Feature DP-SGD, and Feature Projective DP-SGD, reporting COCO-style average precision (AP) and recall (AR) metrics. Unlike MPII, HumanART introduces substantial domain shift with stylized, abstract, and artistically distorted human figures, making it a much more challenging setting for generalization under differential privacy. As such, several configurations, especially those involving large clipping norms (\(C = 1.0\)) or training from scratch without pretraining result in negligible utility, often with AP scores below 1\%, and are omitted from the main tables due to lack of interpretability. In the baseline DP-SGD setup (Table \ref{tab:humanart_dp_}), only the most favorable setting (\(C = 0.01, \epsilon = 0.4\)) yields moderate performance, achieving 40.7 AP. However, once projection is introduced (Table \ref{tab:humanart_dp_projection}), we observe significant gains particularly, for fine-tuned models with \(C = 0.01, \epsilon = 0.8\) achieving 38.7 AP, and improvements also appearing at higher clipping norms. Feature DP (Table \ref{tab:humanart_dp_plus_fdp}) follows a similar trend: without projection, utility is lower overall, peaking at 40.5 AP for \(C = 0.01, \epsilon = 0.8\). The most notable performance is achieved by our feature projective DP method (Table \ref{tab:humanart_dp_plus_fdp_plus_projection}), which achieves strong and stable results across all privacy levels. Specifically, fine-tuning with \(C = 0.01, \epsilon = 0.4\) and \(\epsilon = 0.8\) achieves 51.6 A which is the highest across all methods and settings, while maintaining over 50 AP even at \(\epsilon = 0.2\), demonstrating robustness under strict privacy. Impressively, even models fine-tuned from scratch perform well under our proposed method, reaching 46.0 AP at \(\epsilon = 0.8\), which is a significant jump from the near-zero utility of all other methods trained from scratch. Across the board, feature projective DP demonstrates the most reliable and consistent performance, outperforming both vanilla DP-SGD and projection or FDP variants.

\begin{table}[t]
\centering
\caption{MPII Results: DP-SGD.}

\scriptsize                
\setlength{\tabcolsep}{3pt} 
\renewcommand{\arraystretch}{0.8}  
\resizebox{\columnwidth}{!}{%
\begin{tabular}{l c c c c c c c c c}
\toprule
Privacy Parameter($\epsilon$) & Head & Shoulder & Elbow & Wrist & Hip & Knee & Ankle & Mean & Mean@0.1 \\
\midrule
\multicolumn{10}{c}{\textbf{Finetuning}} \\
\midrule
\multicolumn{10}{l}{\textbf{C = 1.0}} \\
$\epsilon = 0.2$ & 0.00 & 5.15 & 13.31 & 6.00 & 10.20 & 2.18 & 0.21 & 5.94 & 0.29 \\
$\epsilon = 0.4$ & 0.99 & 6.27 & 14.83 & 7.68 & 12.97 & 2.38 & 3.26 & 8.36 & 0.34 \\
$\epsilon = 0.6$ & 0.31 & 13.06 & 21.56 & 8.43 & 19.16 & 3.00 & 3.59 & 12.19 & 0.58 \\
$\epsilon = 0.8$ & 2.97 & 10.43 & 19.16 & 7.06 & 22.31 & 3.75 & 8.12 & 12.53 & 0.61 \\

\midrule
\multicolumn{10}{l}{\textbf{C = 0.1}} \\
$\epsilon = 0.2$ & 30.73 & 35.51 & 31.19 & 13.43 & 37.96 & 12.59 & 13.30 & 28.46 & 1.61 \\
$\epsilon = 0.4$ & 43.01 & 49.92 & 47.98 & 20.20 & 50.67 & 18.22 & 21.56 & 39.78 & 2.43 \\
$\epsilon = 0.6$ & 53.10 & 59.51 & 52.00 & 24.55 & 58.54 & 22.65 & 22.65 & 45.18 & 2.86 \\
$\epsilon = 0.8$ & 64.56 & 64.44 & 53.09 & 28.29 & 60.05 & 30.29 & 28.74 & 49.93 & 3.42 \\

\midrule
\multicolumn{10}{l}{\textbf{C = 0.01}} \\
$\epsilon = 0.2$ & 78.14 & 83.36 & 65.21 & 47.49 & 69.98 & 47.35 & 39.02 & 63.85 & 5.68 \\
$\epsilon = 0.4$ & 83.83 & 88.88 & 77.02 & 63.83 & 74.87 & 62.50 & 52.12 & 73.77 & 9.30 \\
$\epsilon = 0.6$ & 87.11 & 90.05 & 78.71 & 70.05 & 75.37 & 66.61 & 59.42 & 76.85 & 11.05 \\
$\epsilon = 0.8$ & 87.79 & 90.32 & 78.95 & 71.99 & 77.91 & 69.07 & 61.55 & 78.17 & 11.91 \\

\midrule
\multicolumn{10}{c }{\textbf{Finetuning from scratch}} \\
\midrule
\multicolumn{10}{l}{\textbf{C = 1.0}} \\
$\epsilon = 0.2$ & 4.09 & 2.96 & 0.82 & 1.51 & 0.71 & 0.75 & 0.59 & 1.39 & 0.06 \\
$\epsilon = 0.4$ & 6.51 & 7.51 & 3.44 & 3.67 & 1.66 & 14.04 & 1.06 & 5.95 & 0.22 \\
$\epsilon = 0.6$ & 16.17 & 11.79 & 7.41 & 3.92 & 9.23 & 9.27 & 1.82 & 8.67 & 0.37 \\
$\epsilon = 0.8$ & 8.94 & 17.05 & 8.40 & 3.79 & 10.04 & 8.38 & 2.95 & 9.34 & 0.39 \\

\midrule
\multicolumn{10}{l}{\textbf{C = 0.1}} \\
$\epsilon = 0.2$ & 12.48 & 20.87 & 15.70 & 10.91 & 23.39 & 13.96 & 8.36 & 16.11 & 0.68 \\
$\epsilon = 0.4$ & 16.68 & 22.69 & 21.15 & 11.10 & 24.10 & 12.98 & 9.38 & 18.64 & 0.80 \\
$\epsilon = 0.6$ & 23.26 & 28.07 & 21.70 & 13.06 & 26.92 & 14.57 & 9.73 & 21.73 & 1.07 \\
$\epsilon = 0.8$ & 28.58 & 29.14 & 21.36 & 13.18 & 27.40 & 14.31 & 9.09 & 22.74 & 1.10 \\

\midrule
\multicolumn{10}{l}{\textbf{C = 0.01}} \\
$\epsilon = 0.2$ & 28.89 & 31.52 & 22.43 & 12.11 & 27.37 & 16.72 & 10.39 & 24.05 & 1.16 \\
$\epsilon = 0.4$ & 49.25 & 43.19 & 26.28 & 15.01 & 33.58 & 18.05 & 15.28 & 30.94 & 1.86 \\
$\epsilon = 0.6$ & 60.20 & 50.68 & 30.99 & 16.99 & 37.84 & 20.15 & 20.34 & 35.77 & 2.25 \\
$\epsilon = 0.8$ & 62.28 & 54.33 & 36.48 & 21.66 & 43.36 & 22.43 & 21.42 & 39.86 & 2.86 \\

\midrule
\multicolumn{10}{c}{\textbf{Training from scratch}} \\
\midrule
\multicolumn{10}{l}{\textbf{C = 1.0}} \\
$\epsilon = 0.2$ & 0.14 & 0.05 & 0.37 & 0.02 & 0.64 & 0.26 & 0.35 & 0.30 & 0.02 \\
$\epsilon = 0.4$ & 1.71 & 0.07 & 0.07 & 0.15 & 0.28 & 3.28 & 0.02 & 0.68 & 0.03 \\
$\epsilon = 0.6$ & 0.03 & 4.64 & 0.00 & 1.25 & 0.90 & 3.10 & 0.05 & 1.45 & 0.07 \\
$\epsilon = 0.8$ & 0.31 & 0.00 & 1.76 & 0.26 & 0.00 & 0.00 & 0.90 & 0.44 & 0.02 \\

\midrule
\multicolumn{10}{l}{\textbf{C = 0.1}} \\
$\epsilon = 0.2$ & 1.09 & 0.03 & 2.97 & 7.01 & 1.28 & 1.23 & 0.50 & 2.33 & 0.09 \\
$\epsilon = 0.4$ & 0.14 & 5.04 & 6.49 & 4.69 & 18.87 & 0.67 & 0.24 & 5.84 & 0.23 \\
$\epsilon = 0.6$ & 0.10 & 9.51 & 8.01 & 9.58 & 22.90 & 2.08 & 1.23 & 8.12 & 0.33 \\
$\epsilon = 0.8$ & 0.24 & 8.07 & 5.25 & 9.46 & 17.03 & 2.04 & 2.39 & 6.85 & 0.28 \\

\midrule
\multicolumn{10}{l}{\textbf{C = 0.01}} \\
$\epsilon = 0.2$ & 0.31 & 10.31 & 10.07 & 4.90 & 13.33 & 3.41 & 1.77 & 8.17 & 0.36 \\
$\epsilon = 0.4$ & 1.33 & 15.61 & 9.17 & 8.96 & 19.92 & 3.87 & 1.32 & 10.13 & 0.48 \\
$\epsilon = 0.6$ & 6.79 & 17.24 & 13.50 & 9.60 & 21.53 & 5.74 & 2.13 & 12.68 & 0.56 \\
$\epsilon = 0.8$ & 13.27 & 16.76 & 13.38 & 9.68 & 19.87 & 5.48 & 2.17 & 12.74 & 0.54 \\

\bottomrule
\label{tab:mpii_results}
\end{tabular}%
}
\end{table}

\begin{table}[t]
\centering
\caption{MPII Results: DP-SGD with Projection.}
\label{tab:mpii_new_results}
\scriptsize
\setlength{\tabcolsep}{3pt}
\renewcommand{\arraystretch}{0.8}
\resizebox{\columnwidth}{!}{%
\begin{tabular}{l c c c c c c c c c}
\toprule
Privacy Parameter($\epsilon$) & Head & Shoulder & Elbow & Wrist & Hip & Knee & Ankle & Mean & Mean@0.1 \\
\midrule
\multicolumn{10}{c}{\textbf{Finetuning}} \\
\midrule
\multicolumn{10}{l}{\textbf{C = 1.0}} \\
$\epsilon = 0.2$ & 3.48 & 14.79 & 6.85 & 8.38 & 17.36 & 8.85 & 8.62 & 10.34 & 0.42 \\
$\epsilon = 0.4$ & 54.67 & 55.45 & 32.54 & 23.70 & 37.23 & 21.44 & 13.30 & 36.99 & 2.33 \\
$\epsilon = 0.6$ & 55.97 & 49.10 & 31.99 & 28.22 & 39.67 & 24.10 & 14.34 & 37.90 & 2.31 \\
$\epsilon = 0.8$ & 71.18 & 76.19 & 55.51 & 46.60 & 53.97 & 39.13 & 32.97 & 56.26 & 4.19 \\

\midrule
\multicolumn{10}{l}{\textbf{C = 0.1}} \\
$\epsilon = 0.2$ & 88.85 & 89.44 & 75.78 & 68.46 & 61.21 & 63.83 & 54.68 & 73.13 & 10.56 \\
$\epsilon = 0.4$ & 88.44 & 89.84 & 78.92 & 72.62 & 70.21 & 69.45 & 61.08 & 77.17 & 12.55 \\
$\epsilon = 0.6$ & 90.31 & 90.20 & 79.27 & 71.43 & 66.02 & 68.75 & 58.01 & 76.11 & 12.21 \\
$\epsilon = 0.8$ & 91.51 & 90.39 & 79.51 & 72.84 & 71.23 & 67.86 & 59.78 & 77.41 & 12.88 \\

\midrule
\multicolumn{10}{l}{\textbf{C = 0.01}} \\
$\epsilon = 0.2$ & 92.02 & 90.78 & 79.10 & 72.47 & 72.72 & 70.74 & 64.29 & 78.48 & 13.67 \\
$\epsilon = 0.4$ & 91.81 & 90.74 & 79.92 & 72.04 & 75.42 & 71.79 & 65.78 & 79.23 & 13.49 \\
$\epsilon = 0.6$ & 92.29 & 91.78 & 80.48 & 73.75 & 74.16 & 72.29 & 67.88 & 79.89 & 14.28 \\
$\epsilon = 0.8$ & 92.29 & 91.49 & 80.86 & 74.52 & 75.32 & 73.91 & 69.77 & 80.63 & 14.61 \\

\midrule
\multicolumn{10}{c}{\textbf{Finetuning from scratch}} \\
\midrule
\multicolumn{10}{l}{\textbf{C = 1.0}} \\
$\epsilon = 0.2$ & 0.48 & 8.93 & 8.86 & 8.24 & 20.72 & 3.77 & 1.75 & 8.64 & 0.31 \\
$\epsilon = 0.4$ & 4.13 & 14.88 & 14.32 & 6.99 & 3.41 & 5.80 & 3.73 & 8.74 & 0.38 \\
$\epsilon = 0.6$ & 3.48 & 16.34 & 9.90 & 12.35 & 13.21 & 13.48 & 10.23 & 11.85 & 0.56 \\
$\epsilon = 0.8$ & 2.69 & 15.73 & 11.20 & 11.79 & 19.65 & 6.83 & 1.94 & 11.22 & 0.51 \\

\midrule
\multicolumn{10}{l}{\textbf{C = 0.1}} \\
$\epsilon = 0.2$ & 4.40 & 18.99 & 16.99 & 9.42 & 18.07 & 10.70 & 6.78 & 13.58 & 0.63 \\
$\epsilon = 0.4$ & 12.45 & 15.30 & 17.25 & 10.90 & 21.50 & 9.61 & 7.01 & 14.36 & 0.61 \\
$\epsilon = 0.6$ & 15.59 & 15.64 & 16.70 & 10.50 & 19.49 & 14.19 & 7.98 & 15.43 & 0.66 \\
$\epsilon = 0.8$ & 13.34 & 23.30 & 15.51 & 10.08 & 20.06 & 8.68 & 9.05 & 15.92 & 0.66 \\

\midrule
\multicolumn{10}{l}{\textbf{C = 0.01}} \\
$\epsilon = 0.2$ & 82.44 & 69.58 & 49.75 & 43.23 & 43.31 & 39.11 & 36.56 & 53.54 & 5.87 \\
$\epsilon = 0.4$ & 83.77 & 75.70 & 55.19 & 50.44 & 52.12 & 46.12 & 45.35 & 59.82 & 7.87 \\
$\epsilon = 0.6$ & 86.02 & 74.25 & 61.31 & 52.96 & 51.39 & 46.75 & 45.42 & 61.09 & 8.61 \\
$\epsilon = 0.8$ & 87.14 & 77.77 & 63.32 & 56.18 & 57.14 & 50.92 & 48.89 & 64.28 & 9.68 \\

\midrule
\multicolumn{10}{c}{\textbf{Training from scratch}} \\
\midrule
\multicolumn{10}{l}{\textbf{C = 1.0}} \\
$\epsilon = 0.2$ & 0.07 & 1.77 & 10.07 & 11.07 & 12.74 & 1.37 & 0.02 & 5.65 & 0.26 \\
$\epsilon = 0.4$ & 1.36 & 6.98 & 17.69 & 9.77  & 5.24  & 0.95 & 0.07 & 6.76 & 0.28 \\
$\epsilon = 0.6$ & 0.07 & 1.00 & 14.47 & 12.44 & 6.42  & 7.19 & 1.87 & 6.57 & 0.31 \\
$\epsilon = 0.8$ & 0.17 & 7.24 & 11.71 & 5.91  & 2.68  & 4.27 & 1.23 & 5.89 & 0.24 \\

\midrule
\multicolumn{10}{l}{\textbf{C = 0.1}} \\
$\epsilon = 0.2$ & 1.19 & 18.05 & 12.78 & 9.53  & 22.66 & 7.64 & 5.95 & 13.05 & 0.56 \\
$\epsilon = 0.4$ & 0.75 & 13.08 & 6.17  & 5.55  & 21.91 & 3.36 & 5.50 & 9.80  & 0.41 \\
$\epsilon = 0.6$ & 1.98 & 21.28 & 8.16  & 11.05 & 19.70 & 4.11 & 4.72 & 12.24 & 0.56 \\
$\epsilon = 0.8$ & 1.30 & 13.20 & 4.48  & 8.31  & 22.62 & 6.73 & 4.27 & 10.57 & 0.43 \\
\midrule
\multicolumn{10}{l}{\textbf{C = 0.01}} \\
$\epsilon = 0.2$ & 5.15 & 15.10 & 15.07 & 12.39 & 19.61 & 10.28 & 8.83 & 14.26 & 0.65 \\
$\epsilon = 0.4$ & 9.21 & 20.60 & 15.78 & 10.67 & 22.33 & 13.06 & 7.01 & 15.54 & 0.70 \\
$\epsilon = 0.6$ & 17.09 & 19.70 & 6.89  & 10.50 & 22.36 & 15.88 & 10.51 & 15.15 & 0.62 \\
$\epsilon = 0.8$ & 9.48 & 19.55 & 16.12 & 10.91 & 22.45 & 6.61  & 3.57  & 13.96 & 0.60 \\
\bottomrule
\end{tabular}%
}
\end{table}

\begin{table}[t]
\centering
\caption{MPII Results: Feature DP.}
\label{tab:mpii_blur}
\scriptsize                
\setlength{\tabcolsep}{3pt} 
\renewcommand{\arraystretch}{0.8}  
\resizebox{\columnwidth}{!}{%
\begin{tabular}{l c c c c c c c c c}
\toprule
Privacy Parameter($\epsilon$) & Head & Shoulder & Elbow & Wrist & Hip & Knee & Ankle & Mean & Mean@0.1 \\
\midrule
\multicolumn{10}{c}{\textbf{Finetuning}} \\
\midrule
\multicolumn{10}{l}{\textbf{C = 0.01}} \\
$\epsilon = 0.2$ & 87.04 & 89.28 & 75.78 & 66.54 & 75.89 & 65.04 & 58.48 & 75.47 & 10.64 \\
$\epsilon = 0.4$ & 90.76 & 91.30 & 78.51 & 69.95 & 78.55 & 68.85 & 63.86 & 78.60 & 12.71 \\
$\epsilon = 0.6$ & 91.47 & 91.95 & 79.63 & 71.99 & 79.87 & 71.00 & 65.45 & 79.90 & 13.62 \\
$\epsilon = 0.8$ & 92.16 & 92.15 & 79.85 & 72.81 & 80.41 & 71.55 & 66.30 & 80.41 & 14.22 \\
\midrule
\multicolumn{10}{l}{\textbf{C = 0.1}} \\
$\epsilon = 0.2$ & 52.42 & 65.78 & 57.32 & 27.43 & 53.31 & 36.05 & 26.03 & 48.99 & 3.53 \\
$\epsilon = 0.4$ & 78.04 & 81.62 & 64.53 & 40.43 & 67.70 & 45.14 & 39.58 & 61.73 & 5.44 \\
$\epsilon = 0.6$ & 78.48 & 84.80 & 68.48 & 45.60 & 69.88 & 49.89 & 45.11 & 65.32 & 6.47 \\
$\epsilon = 0.8$ & 82.74 & 85.21 & 70.29 & 48.50 & 71.39 & 54.50 & 47.85 & 67.60 & 7.03 \\
\midrule
\multicolumn{10}{l}{\textbf{C = 1.0}} \\
$\epsilon = 0.2$ & 1.60 & 9.90 & 7.86 & 11.57 & 14.25 & 3.00 & 7.91 & 9.32 & 0.33 \\
$\epsilon = 0.4$ & 11.49 & 17.05 & 8.30 & 11.08 & 23.89 & 8.20 & 9.57 & 15.01 & 0.59 \\
$\epsilon = 0.6$ & 14.84 & 25.00 & 12.36 & 13.62 & 27.07 & 10.84 & 11.29 & 18.37 & 0.81 \\
$\epsilon = 0.8$ & 21.69 & 29.28 & 16.16 & 14.56 & 26.55 & 15.66 & 15.45 & 22.05 & 1.01 \\

\midrule
\multicolumn{10}{c}{\textbf{Finetuning from scratch}} \\
\midrule
\multicolumn{10}{l}{\textbf{C = 0.01}} \\
$\epsilon = 0.2$ & 71.15 & 57.17 & 38.71 & 22.32 & 47.24 & 28.39 & 29.12 & 43.89 & 3.48 \\
$\epsilon = 0.4$ & 78.75 & 69.40 & 47.15 & 33.34 & 54.86 & 38.87 & 37.72 & 53.00 & 5.10 \\
$\epsilon = 0.6$ & 82.44 & 72.69 & 52.00 & 38.19 & 59.22 & 41.69 & 41.14 & 56.78 & 6.07 \\
$\epsilon = 0.8$ & 83.80 & 74.81 & 54.88 & 41.29 & 60.93 & 44.35 & 42.80 & 58.98 & 6.70 \\
\midrule
\multicolumn{10}{l}{\textbf{C = 0.1}} \\
$\epsilon = 0.2$ & 33.94 & 30.42 & 19.36 & 12.83 & 27.19 & 15.03 & 10.82 & 23.15 & 1.14 \\
$\epsilon = 0.4$ & 40.86 & 37.75 & 22.57 & 15.08 & 31.71 & 18.09 & 15.94 & 28.46 & 1.53 \\
$\epsilon = 0.6$ & 47.68 & 43.27 & 25.09 & 16.31 & 35.35 & 19.56 & 17.78 & 31.90 & 1.92 \\
$\epsilon = 0.8$ & 52.25 & 45.31 & 27.27 & 16.57 & 35.90 & 19.28 & 18.47 & 33.27 & 2.21 \\
\midrule
\multicolumn{10}{l}{\textbf{C = 1.0}} \\
$\epsilon = 0.2$ & 12.14 & 8.85 & 8.76 & 10.54 & 11.72 & 9.47 & 4.04 & 10.03 & 0.39 \\
$\epsilon = 0.4$ & 10.06 & 13.33 & 10.74 & 9.92 & 20.84 & 12.17 & 5.48 & 12.57 & 0.51 \\
$\epsilon = 0.6$ & 7.03 & 12.65 & 17.11 & 11.29 & 21.67 & 15.35 & 3.19 & 12.84 & 0.52 \\
$\epsilon = 0.8$ & 13.47 & 19.74 & 15.95 & 9.58 & 22.95 & 12.65 & 8.90 & 15.69 & 0.60 \\

\midrule
\multicolumn{10}{c}{\textbf{Training from scratch}} \\
\midrule
\multicolumn{10}{l}{\textbf{C = 0.01}} \\
$\epsilon = 0.2$ & 6.92 & 18.29 & 17.40 & 10.96 & 22.62 & 14.89 & 3.71 & 15.17 & 0.60 \\
$\epsilon = 0.4$ & 10.57 & 22.61 & 18.41 & 12.15 & 23.09 & 14.33 & 6.83 & 17.14 & 0.71 \\
$\epsilon = 0.6$ & 11.39 & 22.18 & 17.16 & 11.91 & 24.23 & 15.90 & 8.86 & 17.57 & 0.82 \\
$\epsilon = 0.8$ & 14.09 & 21.31 & 19.52 & 11.31 & 24.32 & 16.44 & 7.35 & 17.74 & 0.80 \\
\midrule
\multicolumn{10}{l}{\textbf{C = 0.1}} \\
$\epsilon = 0.2$ & 4.23 & 0.56 & 7.82 & 8.64 & 19.44 & 1.91 & 1.96 & 7.13 & 0.26 \\
$\epsilon = 0.4$ & 0.48 & 12.62 & 10.69 & 8.12 & 21.15 & 2.36 & 0.33 & 8.87 & 0.37 \\
$\epsilon = 0.6$ & 0.48 & 15.08 & 10.64 & 9.00 & 19.72 & 4.33 & 1.20 & 9.60 & 0.41 \\
$\epsilon = 0.8$ & 0.61 & 13.09 & 12.94 & 10.74 & 22.73 & 4.94 & 1.37 & 11.22 & 0.48 \\
\midrule
\multicolumn{10}{l}{\textbf{C = 1.0}} \\
$\epsilon = 0.2$ & 0.31 & 0.32 & 0.14 & 0.00 & 0.03 & 0.06 & 0.12 & 0.35 & 0.02 \\
$\epsilon = 0.4$ & 1.84 & 3.63 & 0.02 & 0.43 & 0.00 & 0.00 & 0.09 & 0.78 & 0.02 \\
$\epsilon = 0.6$ & 0.00 & 2.62 & 0.05 & 4.61 & 8.34 & 0.12 & 0.00 & 2.39 & 0.10 \\
$\epsilon = 0.8$ & 0.14 & 0.02 & 0.07 & 0.00 & 15.22 & 0.02 & 0.26 & 2.57 & 0.12 \\

\bottomrule
\end{tabular}%
}
\end{table}

\begin{table}[t]
\centering
\caption{MPII Results: Feature Projective DP.}
\label{tab:mpii_blur_projection}
\scriptsize                
\setlength{\tabcolsep}{3pt} 
\renewcommand{\arraystretch}{0.8}  
\resizebox{\columnwidth}{!}{%
\begin{tabular}{l c c c c c c c c c}
\toprule
Privacy Parameter($\epsilon$) & Head & Shoulder & Elbow & Wrist & Hip & Knee & Ankle & Mean & Mean@0.1 \\
\midrule
\multicolumn{10}{c}{\textbf{Finetuning}} \\
\midrule
\multicolumn{10}{l}{\textbf{C = 0.01}} \\
$\epsilon = 0.2$ & 94.17 & 92.70 & 82.17 & 73.29 & 80.79 & 76.45 & 72.60 & 82.50 & 19.65 \\
$\epsilon = 0.4$ & 94.13 & 92.70 & 81.46 & 73.10 & 80.51 & 75.44 & 71.00 & 82.01 & 19.23 \\
$\epsilon = 0.6$ & 94.10 & 92.70 & 81.47 & 72.79 & 80.87 & 75.32 & 71.19 & 82.01 & 19.48 \\
$\epsilon = 0.8$ & 94.27 & 92.63 & 81.58 & 72.50 & 80.37 & 75.56 & 70.88 & 81.91 & 19.22 \\
\midrule
\multicolumn{10}{l}{\textbf{C = 0.1}} \\
$\epsilon = 0.2$ & 93.79 & 91.85 & 79.07 & 71.56 & 77.43 & 73.32 & 68.92 & 80.24 & 15.20 \\
$\epsilon = 0.4$ & 94.03 & 92.56 & 80.72 & 73.19 & 79.85 & 74.95 & 69.96 & 81.60 & 17.03 \\
$\epsilon = 0.6$ & 93.86 & 92.82 & 81.17 & 74.70 & 79.78 & 75.68 & 70.78 & 82.09 & 17.74 \\
$\epsilon = 0.8$ & 94.78 & 93.21 & 82.31 & 74.51 & 80.63 & 76.04 & 71.30 & 82.62 & 18.64 \\
\midrule
\multicolumn{10}{l}{\textbf{C = 1.0}} \\
$\epsilon = 0.2$ & 73.36 & 65.88 & 53.72 & 40.79 & 56.14 & 43.44 & 34.10 & 54.75 & 4.05 \\
$\epsilon = 0.4$ & 86.53 & 82.17 & 63.52 & 54.50 & 57.68 & 49.20 & 43.01 & 64.02 & 6.09 \\
$\epsilon = 0.6$ & 86.66 & 86.06 & 66.70 & 51.94 & 67.47 & 51.40 & 47.02 & 67.02 & 6.73 \\
$\epsilon = 0.8$ & 91.13 & 89.06 & 73.91 & 63.39 & 62.92 & 59.60 & 51.75 & 71.66 & 9.37 \\

\midrule
\multicolumn{10}{c}{\textbf{Finetuning from scratch}} \\
\midrule
\multicolumn{10}{l}{\textbf{C = 0.01}} \\
$\epsilon = 0.2$ & 92.91 & 88.94 & 76.55 & 67.33 & 77.81 & 71.11 & 65.99 & 78.11 & 16.82 \\
$\epsilon = 0.4$ & 92.74 & 88.55 & 75.80 & 65.63 & 77.48 & 70.02 & 65.37 & 77.41 & 16.23 \\
$\epsilon = 0.6$ & 93.21 & 88.62 & 75.61 & 65.15 & 77.79 & 69.33 & 64.52 & 77.23 & 16.37 \\
$\epsilon = 0.8$ & 92.29 & 87.11 & 74.11 & 63.13 & 76.53 & 67.46 & 62.78 & 75.74 & 15.83 \\
\midrule
\multicolumn{10}{l}{\textbf{C = 0.1}} \\
$\epsilon = 0.2$ & 91.95 & 85.61 & 71.11 & 59.78 & 73.84 & 64.58 & 61.36 & 73.51 & 14.33 \\
$\epsilon = 0.4$ & 93.76 & 88.65 & 76.97 & 67.07 & 77.27 & 70.30 & 66.23 & 78.01 & 16.93 \\
$\epsilon = 0.6$ & 93.49 & 89.08 & 77.09 & 67.81 & 79.09 & 72.23 & 67.48 & 78.86 & 16.93 \\
$\epsilon = 0.8$ & 94.03 & 90.46 & 78.56 & 68.58 & 80.68 & 72.54 & 69.08 & 79.95 & 17.83 \\
\midrule
\multicolumn{10}{l}{\textbf{C = 1.0}} \\
$\epsilon = 0.2$ & 4.20 & 8.02 & 13.07 & 9.20 & 22.99 & 7.64 & 7.02 & 10.98 & 0.46 \\
$\epsilon = 0.4$ & 10.54 & 17.53 & 13.19 & 11.50 & 20.60 & 7.41 & 5.12 & 13.49 & 0.53 \\
$\epsilon = 0.6$ & 15.52 & 20.67 & 14.91 & 12.11 & 21.64 & 14.99 & 8.10 & 16.29 & 0.78 \\
$\epsilon = 0.8$ & 20.36 & 20.92 & 12.63 & 10.95 & 24.98 & 10.07 & 8.46 & 16.29 & 0.73 \\

\midrule
\multicolumn{10}{c}{\textbf{Training from scratch}} \\
\midrule
\multicolumn{10}{l}{\textbf{C = 0.01}} \\
$\epsilon = 0.2$ & 16.75 & 19.53 & 17.62 & 12.41 & 24.84 & 14.35 & 9.99 & 17.34 & 0.71 \\
$\epsilon = 0.4$ & 62.65 & 52.11 & 34.07 & 19.03 & 41.09 & 27.30 & 23.74 & 38.90 & 3.43 \\
$\epsilon = 0.6$ & 71.66 & 60.36 & 38.86 & 22.51 & 46.65 & 31.51 & 26.26 & 44.28 & 4.39 \\
$\epsilon = 0.8$ & 67.84 & 58.95 & 38.06 & 21.86 & 44.95 & 31.99 & 27.66 & 43.39 & 4.12 \\
\midrule
\multicolumn{10}{l}{\textbf{C = 0.1}} \\
$\epsilon = 0.2$ & 14.02 & 17.56 & 14.69 & 10.86 & 20.65 & 15.84 & 7.96 & 15.50 & 0.72 \\
$\epsilon = 0.4$ & 11.02 & 19.51 & 16.62 & 11.02 & 23.61 & 14.16 & 10.16 & 16.42 & 0.65 \\
$\epsilon = 0.6$ & 18.21 & 24.56 & 19.24 & 12.34 & 26.62 & 15.68 & 12.49 & 19.78 & 0.93 \\
$\epsilon = 0.8$ & 53.27 & 46.01 & 29.37 & 17.27 & 35.47 & 21.20 & 17.17 & 33.49 & 2.33 \\
\midrule
\multicolumn{10}{l}{\textbf{C = 1.0}} \\
$\epsilon = 0.2$ & 1.13 & 14.74 & 9.56 & 8.19 & 22.14 & 14.47 & 3.54 & 12.39 & 0.54 \\
$\epsilon = 0.4$ & 17.77 & 14.81 & 15.85 & 9.46 & 22.16 & 14.83 & 4.06 & 15.23 & 0.70 \\
$\epsilon = 0.6$ & 18.49 & 21.08 & 15.39 & 11.12 & 22.54 & 5.14 & 5.48 & 14.67 & 0.66 \\
$\epsilon = 0.8$ & 5.83 & 14.96 & 14.88 & 10.66 & 22.21 & 13.04 & 3.83 & 12.97 & 0.55 \\

\bottomrule
\end{tabular}%
}
\end{table}

\begin{table}[t]
\centering
\caption{HumanART Results: DP-SGD.}
\label{tab:humanart_dp_}
\scriptsize                
\setlength{\tabcolsep}{3pt} 
\renewcommand{\arraystretch}{0.8}  
\resizebox{\columnwidth}{!}{%
\begin{tabular}{l c c c c c c c c c c}
\toprule
Privacy Parameter($\epsilon$) & $AP$ & $AP^{50}$ & $AP^{75}$ & $AP^M$ & $AP^L$ & $AR$ & $AR^{50}$ & $AR^{75}$ & $AR^M$ & $AR^L$ \\
\midrule
\multicolumn{10}{c}{\textbf{Finetuning}} \\
\midrule
\multicolumn{10}{l}{\textbf{C = 0.01}} \\
$\epsilon = 0.2$ & 29.1& 67.6 &20.0  &14.0  & 30.9 & 34.0 & 71.6 &28.6  &22.0  &35.5\\
$\epsilon = 0.4$ &40.7 & 74.7 &39.3  &23.8  & 42.7 & 45.8 & 77.1 & 46.7 &32.9  &47.9\\
$\epsilon = 0.6$ &37.5 &74.7  &33.6  &20.9  &39.4  &42.1  &77.6  &41.4  &29.1  &43.8\\
$\epsilon = 0.8$ &39.0 & 75.9 &36.0  &22.2  &40.9  &43.5  &78.3  &43.6  &30.1  &45.3\\
\midrule
\multicolumn{10}{l}{\textbf{C = 0.1}} \\
$\epsilon = 0.2$ & 3.5& 19.0 &0.0  &0.9  &4.0  &8.7  &35.6  & 0.8 & 5.0 &9.2\\
$\epsilon = 0.4$ & 7.7& 32.4 &0.5  &2.3  &8.5  &14.6  &47.7  &4.0  &8.7  &15.3\\
$\epsilon = 0.6$ & 10.4& 39.3 &1.5  &3.4  &11.3  &17.1  &51.9  &6.2  &10.5  &17.9\\
$\epsilon = 0.8$ & 12.0& 43.3 &2.4  &3.9  &13.0  &18.4  &54.1  &7.3  &11.4  &19.2\\


\midrule
\multicolumn{10}{c}{\textbf{Finetuning from scratch}} \\
\midrule
\multicolumn{10}{l}{\textbf{C = 0.01}} \\
$\epsilon = 0.2$ &0.9 &5.8  &0.0  &0.2  &1.0  &3.8  &19.7  &0.1  &2.6  &3.9\\
$\epsilon = 0.4$ & 1.3& 8.0 &0.0  &0.4  &1.4  &5.0  &23.7  &0.3  &3.5  &5.2\\
$\epsilon = 0.6$ & 1.7& 10.7 &0.0  &0.5  &1.9  &6.1  &27.5  &0.6  &4.2  &6.4\\
$\epsilon = 0.8$ & 2.1& 12.9 &0.0  &0.8  &2.3  &6.8  &29.6  &0.7  &5.1  &7.0\\
\midrule
\multicolumn{10}{l}{\textbf{C = 0.1}} \\
$\epsilon = 0.2$ & 0.0& 0.4 &0.0  &0.0  &0.1  &0.8  &5.7  &0.0  &0.5  &0.8\\
$\epsilon = 0.4$ & 0.2& 1.9 &0.0  &0.1  &0.3  &2.3  &13.4  &0.1  &1.4  &2.4\\
$\epsilon = 0.6$ & 0.5& 3.6 &0.0  &0.1  &0.6  &3.3  &18.6  &0.0  &2.2  &3.4\\
$\epsilon = 0.8$ & 0.6& 4.3 &0.0  &0.1  &0.7  &3.5  & 20.0 & 0.0 &2.4  &3.7\\


\bottomrule
\end{tabular}%
}
\end{table}

\begin{table}[t]
\centering
\caption{HumanART Results: DP-SGD with projection.}
\label{tab:humanart_dp_projection}
\scriptsize                
\setlength{\tabcolsep}{3pt} 
\renewcommand{\arraystretch}{0.8}  
\resizebox{\columnwidth}{!}{%
\begin{tabular}{l c c c c c c c c c c}
\toprule
Privacy Parameter($\epsilon$) & $AP$ & $AP^{50}$ & $AP^{75}$ & $AP^M$ & $AP^L$ & $AR$ & $AR^{50}$ & $AR^{75}$ & $AR^M$ & $AR^L$ \\
\midrule
\multicolumn{10}{c}{\textbf{Finetuning}} \\
\midrule
\multicolumn{10}{l}{\textbf{C = 0.01}} \\
$\epsilon = 0.2$ &38.3 & 73.9 & 35.6 & 21.7 & 40.4 &43.6  &76.9  &43.6  &30.8  &45.8\\
$\epsilon = 0.4$ & 34.7& 73.4 & 29.1 & 18.3 & 36.6 & 39.7 & 76.0 & 37.9 & 26.9 &41.4\\
$\epsilon = 0.6$ & 40.3& 75.8 & 38.4 & 22.5 & 42.6 & 46.1 & 78.5 & 47.0 & 33.1 &47.8\\
$\epsilon = 0.8$ & 38.7& 74.3 & 36.3 & 21.1 & 40.8 & 44.3 & 77.3 & 44.5 & 32.1 &45.9\\
\midrule
\multicolumn{10}{l}{\textbf{C = 0.1}} \\
$\epsilon = 0.2$ &31.6 & 69.1 & 24.7 & 17.0 & 33.3 & 36.8 & 72.7 & 33.1 &26.0  &38.2\\
$\epsilon = 0.4$ & 32.9& 70.4 & 27.0 & 18.0 & 34.6 & 38.1 & 73.2 & 35.2 &27.2  &39.5\\
$\epsilon = 0.6$ & 33.6& 71.3 & 26.9 & 18.4 & 35.4 & 38.6 & 74.3 & 35.2 & 28.2 &40.0\\
$\epsilon = 0.8$ & 33.5& 69.4 &28.4  &18.4  &35.3  & 38.7 & 72.6 & 36.8 &27.1  &40.2\\

\midrule
\multicolumn{10}{l}{\textbf{C = 1.0}} \\
$\epsilon = 0.2$ &1.1 &6.7  &0.0  &0.3  &1.3  &4.8  &21.9  &0.2  &3.2  &5.0\\
$\epsilon = 0.4$ & 11.7&  41.7& 2.0 & 3.8 & 12.8 &18.1  &52.4  &7.8  &11.1  &19.0\\
$\epsilon = 0.6$ & 10.7& 39.4 &1.9  &2.9  &11.7  &17.1  &51.2  &6.8  &9.7  &18.0\\
$\epsilon = 0.8$ & 23.5& 60.4 & 14.2 & 11.1 &25.2  &30.2  & 66.2 & 24.2 & 20.2 &31.5\\

\midrule
\multicolumn{10}{c}{\textbf{Finetuning from scratch}} \\
\midrule
\multicolumn{10}{l}{\textbf{C = 0.01}} \\
$\epsilon = 0.2$ &1.7 &8.5  &0.4  & 0.4 & 1.9 & 4.9 & 19.5 & 1.1 & 2.9 &5.2\\
$\epsilon = 0.4$ & 3.5& 15.2 & 0.3 & 1.1 & 3.9 & 8.4 & 29.0 & 2.6 & 5.8 &8.7\\
$\epsilon = 0.6$ & 2.1& 10.2 & 0.1 & 0.7 & 2.4 & 5.9 & 22.2 & 1.4 & 3.4 &6.2\\
$\epsilon = 0.8$ & 1.4& 7.3 & 0.1 & 0.5 & 1.6 & 4.3 & 18.2 & 0.6 & 2.4 &4.5\\
\midrule
\multicolumn{10}{l}{\textbf{C = 0.1}} \\
$\epsilon = 0.2$ & 0.5& 2.8 & 0.1 & 0.2 & 0.6 & 2.2 & 10.9 & 0.1 & 1.5 &2.3\\
$\epsilon = 0.4$ & 0.6&4.0  &0.0  &0.3  &0.7  &2.5  &12.9  &0.1  &1.7  &2.6\\
$\epsilon = 0.6$ & 1.2& 5.7 & 0.5 & 0.4 & 1.5 & 3.4 & 15.3 & 0.6 & 1.7 &3.7\\
$\epsilon = 0.8$ & 1.1& 4.2 & 1.0 & 0.3 & 1.2 & 3.1 & 14.0 & 0.5 & 1.8 &3.2\\

\midrule
\multicolumn{10}{l}{\textbf{C = 1.0}} \\
$\epsilon = 0.2$ &0.1 &0.9  &0.0  &0.0  &0.2  & 0.9 &5.6  &  0.0& 0.6 &0.9\\
$\epsilon = 0.4$ &0.3 &2.0  & 0.1 & 0.0 &0.4  &1.9  &11.1  & 0.1 & 1.0 &2.0\\
$\epsilon = 0.6$ &0.8 & 4.6 &0.0  &0.1  &0.9  &3.1  &16.1  &0.1  &1.7  &3.3\\
$\epsilon = 0.8$ & 0.2& 1.3 & 0.1 & 0.0 & 0.3 & 1.3 & 7.5 & 0.1 & 0.9 &1.4\\

\midrule
\multicolumn{10}{c}{\textbf{Training from scratch}} \\
\midrule
\multicolumn{10}{l}{\textbf{C = 0.01}} \\
$\epsilon = 0.2$ &0.12 &0.68  &0.06  &0.0  &0.17 &1.06  &5.7  &0.1  &0.5&1.1\\
$\epsilon = 0.4$ &0.0 &0.09  &0.0  &0.0  &0.0  &0.09  &0.71  &0.0  &0.10  &0.09\\
$\epsilon = 0.6$ & 0.31& 2.1 & 0.0 & 0.07 & 0.4 & 1.8 & 11.7 & 0.01 &1.18  &1.93\\
$\epsilon = 0.8$ & 0.11& 0.82 &0.0  &0.04  & 0.2 &1.2  & 7.0 &0.02  &1.15  &1.20\\
\midrule
\multicolumn{10}{l}{\textbf{C = 0.1}} \\
$\epsilon = 0.2$ &0.05 &0.29  &0.0  &0.06  &0.38  &2.5  &0.0  &0.3  &0.4  &1.24\\
$\epsilon = 0.4$ &0.12 & 0.91 & 0.0 & 0.05 & 0.16 &1.35  &7.58  &0.04  &1.3  &1.37\\
$\epsilon = 0.6$ & 0.19& 1.27 &0.0  &0.04  & 0.23 & 1.6 & 9.06 & 0.03 & 1.02 &1.6\\
$\epsilon = 0.8$ & 0.02& 0.11 &0.0  &0.0  &0.03  &2.09  &0.0  &0.25  & 0.32 &0.31\\


\bottomrule
\end{tabular}%
}
\end{table}

\begin{table}[t]
\centering
\caption{HumanART Results: Feature Projection DP-SGD.}
\label{tab:humanart_dp_plus_fdp}
\scriptsize                
\setlength{\tabcolsep}{3pt} 
\renewcommand{\arraystretch}{0.8}  
\resizebox{\columnwidth}{!}{%
\begin{tabular}{l c c c c c c c c c c}
\toprule
Privacy Parameter($\epsilon$) & $AP$ & $AP^{50}$ & $AP^{75}$ & $AP^M$ & $AP^L$ & $AR$ & $AR^{50}$ & $AR^{75}$ & $AR^M$ & $AR^L$ \\
\midrule
\multicolumn{10}{c}{\textbf{Finetuning}} \\
\midrule
\multicolumn{10}{l}{\textbf{C = 0.01}} \\
$\epsilon = 0.2$ & 32.6& 71.8 &26.4  &17.1  &34.6  &38.1  &75.2  &35.1  &26.6 &39.6  \\
$\epsilon = 0.4$ & 37.3& 74.6 &34.2  &20.7  &39.5  &42.6  &77.7  &42.4  &30.4 & 44.2  \\
$\epsilon = 0.6$ &39.3 &75.9  &37.6  &22.2  &41.7  &44.5  &78.8  &45.4  &31.4&46.2  \\
$\epsilon = 0.8$ & 40.5& 76.9 &  39.4&23.1  & 42.7 & 45.6 & 79.4 & 47.0 & 32.5 & 47.3\\
\midrule
\multicolumn{10}{l}{\textbf{C = 0.1}} \\
$\epsilon = 0.2$ &8.7 &37.0  &1.1  &3.1  &9.4  &14.1  &48.7  &2.9  &9.6  &14.7\\
$\epsilon = 0.4$ & 13.5& 47.2 &2.1  &4.8  &14.6  &18.8  &55.9  &6.8  &12.3  &19.6\\
$\epsilon = 0.6$ &15.8 & 51.9 &3.5  &5.9  &17.0  &21.1  &59.1  &9.3  &14.1  &22.0\\
$\epsilon = 0.8$ & 17.1& 53.5 &4.6  &6.6  &18.4  &22.4  &60.6  &10.8  &14.9  &23.3\\


\midrule
\multicolumn{10}{c}{\textbf{Finetuning from scratch}} \\
\midrule
\multicolumn{10}{l}{\textbf{C = 0.01}} \\
$\epsilon = 0.2$ & 4.8& 23.0 &0.2  &0.9  &5.5  &10.4  & 38.4 &2.4  &6.3  &10.9\\
$\epsilon = 0.4$ & 7.7& 30.0 &1.7  &1.7  &8.5  &13.6  &43.1  &5.0  &8.5  &14.2\\
$\epsilon = 0.6$ & 8.8& 32.8 &2.3  &2.0  &9.7  &14.8  &45.2  &6.3  &9.3  &15.4\\
$\epsilon = 0.8$ & 9.5& 34.5 &2.6  &2.2  &10.6  &15.7  &46.9  &7.1  &10.1  &16.3\\
\midrule
\multicolumn{10}{l}{\textbf{C = 0.1}} \\
$\epsilon = 0.2$ &0.2 &1.9  &0.0  &0.0  &0.3  &2.3  &13.6  &0.0  &1.4  &2.5\\
$\epsilon = 0.4$ & 0.7& 4.9 &0.0  &0.0  &0.8  &3.2  &18.3  &0.09  &1.7  &3.3\\
$\epsilon = 0.6$ & 1.0& 7.0 &0.0  &0.2  &1.1  &3.9  &21.0  &0.1  &2.7  &4.0\\
$\epsilon = 0.8$ & 1.3& 9.3 &0.0  &0.2  &1.5  &4.4  &23.1  &0.1  &2.8  &4.6\\


\bottomrule
\end{tabular}%
}
\end{table}

\begin{table}[t]
\centering
\caption{HumanART Results: Feature Projection DP-SGD plus projection.}
\label{tab:humanart_dp_plus_fdp_plus_projection}
\scriptsize                
\setlength{\tabcolsep}{3pt} 
\renewcommand{\arraystretch}{0.8}  
\resizebox{\columnwidth}{!}{%
\begin{tabular}{l c c c c c c c c c c}
\toprule
Privacy Parameter($\epsilon$) & $AP$ & $AP^{50}$ & $AP^{75}$ & $AP^M$ & $AP^L$ & $AR$ & $AR^{50}$ & $AR^{75}$ & $AR^M$ & $AR^L$ \\
\midrule
\multicolumn{10}{c}{\textbf{Finetuning}} \\
\midrule
\multicolumn{10}{l}{\textbf{C = 0.01}} \\
$\epsilon = 0.2$ &50.6 &80.2  &53.4  & 31.3 & 52.8 &55.3  &82.4  &59.1  &41.3  &57.2\\
$\epsilon = 0.4$ & 51.6& 80.4 &55.0  &33.5  &53.8  &56.2  &82.7  &60.3  &42.2  &58.1\\
$\epsilon = 0.6$ & 50.9& 80.2 & 53.6 & 32.5 & 53.2 &55.9  &82.7  &59.9  &42.1  &57.8\\
$\epsilon = 0.8$ &51.6 &80.2  &54.7  &33.6  &53.7  &56.2& 82.5 & 60.2 &42.3  &58.1\\
\midrule
\multicolumn{10}{l}{\textbf{C = 0.1}} \\
$\epsilon = 0.2$ &40.4 &75.5  &38.9  &24.4  &42.3  &45.4  &78.0  &46.0  &33.0  &47.1\\
$\epsilon = 0.4$ & 41.1& 76.8 & 40.2 & 24.6 &43.3  &46.8  &79.5  &48.2  &34.4  &48.4\\
$\epsilon = 0.6$ & 42.2& 75.7 & 42.2 &25.2  &44.3  &47.5  &78.6  &49.3  &34.4  &49.2\\
$\epsilon = 0.8$ & 43.0&  77.0& 42.9 &25.7  &45.1  &48.4  &79.9  &50.3  &35.7  &50.1\\

\midrule
\multicolumn{10}{l}{\textbf{C = 1.0}} \\
$\epsilon = 0.2$ & 26.8& 62.1 &19.0  &13.5  &28.4  &32.9  &67.0  &28.6  &23.2  &34.2\\
$\epsilon = 0.4$ &32.4 &70.5  &25.4  &19.0  &34.2  &39.0  &74.7  &36.6  &29.9  &40.3\\
$\epsilon = 0.6$ & 33.9& 70.6 &28.7  &18.3  &35.8  &40.1  &74.4  &38.5  &28.8  &41.6\\
$\epsilon = 0.8$ &35.4 & 72.1 &30.7  &20.2  &37.2  &40.7  &75.1  &38.6  &29.9  &42.1\\

\midrule
\multicolumn{10}{c}{\textbf{Finetuning from scratch}} \\
\midrule
\multicolumn{10}{l}{\textbf{C = 0.01}} \\
$\epsilon = 0.2$ & 45.3& 76.9 &46.4  &29.5  &47.3  &50.4  &79.7  &53.0  &38.0  &52.1\\
$\epsilon = 0.4$ &45.1 &76.6  &46.5  &28.0  &47.2  &50.5  &79.1  &53.5  &36.8  &52.3\\
$\epsilon = 0.6$ &43.8 &75.2  &44.5  &28.5  &45.6  &49.6  &78.0  &52.2  &38.3  &51.1\\
$\epsilon = 0.8$ & 46.0& 76.7 & 47.5 & 29.3 &48.1  &51.4  &79.7  &54.5  &38.0  &53.2\\
\midrule
\multicolumn{10}{l}{\textbf{C = 0.1}} \\
$\epsilon = 0.2$ & 6.8& 24.4 &2.3  &2.8  &7.4  &11.6  &35.9  &5.2  &8.8  &12.0\\
$\epsilon = 0.4$ & 22.0& 53.7 & 14.5 & 12.0 &23.3  &27.7  &60.3  &22.1  &20.6  &28.7\\
$\epsilon = 0.6$ & 29.1& 62.4 & 24.0 & 17.7 &30.8  &34.9  &67.8  &32.0  &26.2  &36.1\\
$\epsilon = 0.8$ & 33.0& 68.0 & 27.9 & 21.5 & 34.4 & 38.6 & 72.0 & 36.5 &30.0  &39.8\\


\bottomrule
\label{tab:humanart}
\end{tabular}%
}
\end{table}

\clearpage

\begin{figure*}[!htb]
\centering
\includegraphics[width=1.0\textwidth]{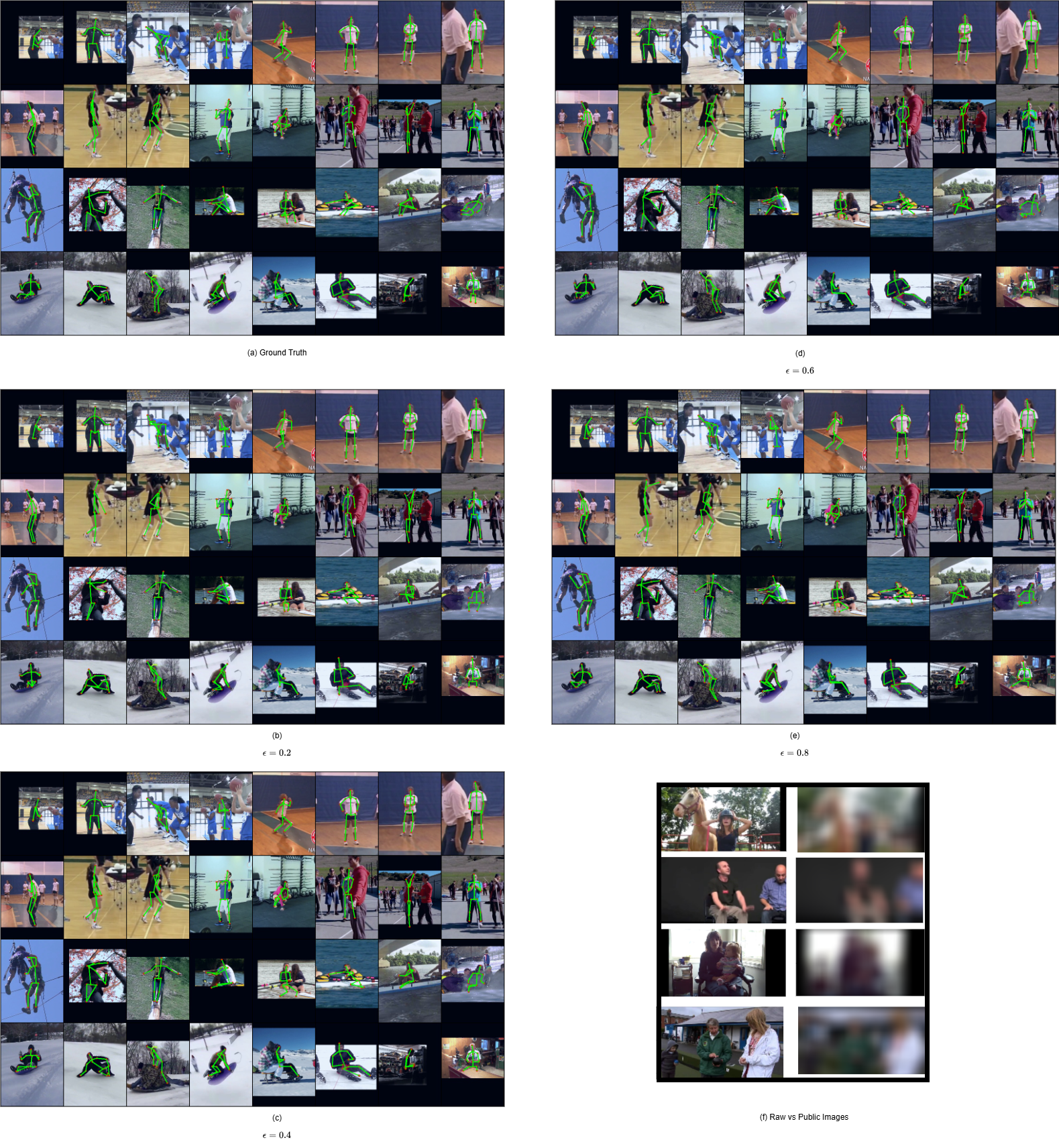}
\caption{Figures (a-e)Depiction of qualitative results on DP-SGD, Projection DP-SGD and Feature Projection DP-SGD. We specifically show results on Finetuning with $C=0.1$ at various privacy budgets. (f) Representation of Raw (Private) image compared to public feature (gaussian blurred).}
\label{fig:qual}
\end{figure*}

\clearpage

\end{document}